\documentclass[letterpaper]{article} 
\usepackage{aaai2026}  
\usepackage{times}  
\usepackage{helvet}  
\usepackage{courier}  
\usepackage[hyphens]{url}  
\usepackage{graphicx} 
\usepackage{natbib}  
\usepackage{caption} 
\usepackage{algorithm}
\usepackage{algorithmic}

\usepackage{newfloat}
\usepackage{listings}

\usepackage{enumitem}
\usepackage{booktabs} 
\usepackage{caption}  
\usepackage{amsmath}
\usepackage{amsfonts}

\usepackage{amssymb}
\usepackage{bm}
\usepackage{cite}
\usepackage{mathtools}
\usepackage{graphicx}
\usepackage{textcomp}
\usepackage{xcolor}
\usepackage{subfigure}
\usepackage{multirow}
\usepackage{colortbl}
\usepackage{float}
\usepackage{array}
\usepackage{newtxmath}
\usepackage{makecell} 
\usepackage{multicol}

\DeclareCaptionStyle{ruled}{labelfont=normalfont,labelsep=colon,strut=off} 
\floatstyle{ruled}
\newfloat{listing}{tb}{lst}{}
\floatname{listing}{Listing}
\title{Keep on Going: Learning Robust Humanoid Motion Skills via Selective Adversarial Training}
\author{
    Yang Zhang\textsuperscript{\rm 1},
    Zhanxiang Cao\textsuperscript{\rm 1,2},
    Buqing Nie\textsuperscript{\rm 1},
    Haoyang Li\textsuperscript{\rm 1,2},
    Zhong Jiangwei\textsuperscript{\rm 3},
    Qiao Sun\textsuperscript{\rm 3},
    Xiaoyi Hu\textsuperscript{\rm 3},
    Xiaokang Yang\textsuperscript{\rm 1},
    Yue Gao\textsuperscript{\rm 1,2}\thanks{Corresponding author.}
}
\affiliations{
    \textsuperscript{\rm 1}MoE Key Lab of Artificial Intelligence and AI Institute, Shanghai Jiao Tong University\\
    \textsuperscript{\rm 2}Shanghai Innovation Institute, Shanghai, China\\
    \textsuperscript{\rm 3}Lenovo Research\\

    \{zhangyang-sjtu-2022, yuegao\}@sjtu.edu.cn
}

\usepackage{bibentry}

\begin{document}

\maketitle

\begin{abstract}
Humanoid robots are expected to operate reliably over long horizons while executing versatile whole-body skills.
Yet Reinforcement Learning (RL) motion policies typically lose stability under prolonged operation, sensor/actuator noise, and real world disturbances.
In this work, we propose a \textbf{S}elective \textbf{A}dversarial \textbf{A}ttack for \textbf{R}obust \textbf{T}raining (\textbf{SA2RT}) to enhance the robustness of motion skills.
The adversary is learned to identify and sparsely perturb the most vulnerable states and actions under an attack-budget constraint, thereby exposing true weakness without inducing conservative overfitting.
The resulting non-zero sum, alternating optimization continually strengthens the motion policy against the strongest discovered attacks.
We validate our approach on the Unitree G1 humanoid robot across perceptive locomotion and whole-body control tasks.
Experimental results show that adversarially trained policies improve the terrain traversal success rate by $40\%$, reduce the trajectory tracking error by $32\%$, and maintain long horizon mobility and tracking performance.
Together, these results demonstrate that selective adversarial attacks are an effective driver for learning robust, long horizon humanoid motion skills.
\end{abstract}


\section{Introduction}\label{Introduction}
Humanoid robots, with their human-like morphology and versatile mobility~\cite{gu2025humanoid}, have the potential to replace humans performing various tasks in daily life~\cite{tong2024advancements}, demonstrating significant potential in fields such as domestic service~\cite{mende2019service}, industrial production~\cite{malik2023intelligent}, and healthcare~\cite{mukherjee2022humanoid}.
Recent advances in Reinforcement Learning (RL)-based motion control enable robots to autonomously learn optimal policies through simulated environmental interactions, achieving complex motion skills~\cite{long2024learning, chen2024learning, van2024revisiting, ji2024exbody2,radosavovic2024humanoid}.
However, due to the differences between the simulation environment and the real world~\cite{zhang2024whole}, including environmental variations~\cite{fujimoto2024assessing}, sensor noise~\cite{liang2022efficient}, and external disturbances~\cite{peng2018sim}, neural controllers encounter domain distribution shift~\cite{fujimoto2024assessing}, leading to the sim-to-real transfer problem of motion policies~\cite{moos2022robust}.
In addition, the inherent sensitivity of RL-based neural controllers to perturbations~\cite{shi2025adversarial} and the lack of systematic robustness design~\cite{barbara2024robust} further exacerbate the instability of deploying these policies in real-world applications~\cite{kobayashi2022l2c2}, resulting in the inability of humanoid robots to complete long horizon motion tasks in complex environments.

\begin{figure}[t]
\centerline{\includegraphics[width=8.2cm]{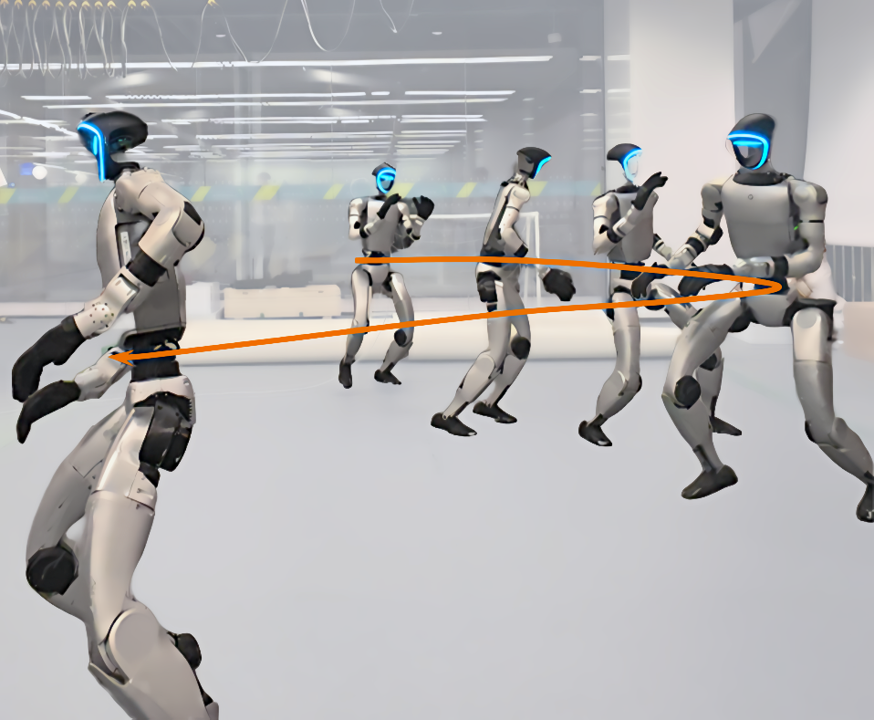}}
\caption{Snapshots of the humanoid robot executing whole-body trajectory tracking.
WBC-SAP can track challenging dynamic trajectories over an extended duration, demonstrating that the SA2RT significantly improves the robustness of motion policies.}
\label{fig1}
\end{figure}

To enhance the robustness of motion policies, previous works employ Domain Randomization (DR) to introduce perturbations~\cite{wei2023learning} such as observation noise and environmental variations during training to simulate the distribution gap between training and deployment, improving the robustness of policies to environmental uncertainty~\cite{gu2024humanoid}.
Other works utilize regularization constraints to indirectly improve the robustness of motion policies~\cite{chen2024learning, yan2024learning}.
Lipschitz regularization limits the output variation of the policy under small input perturbations, improving action smoothness and robustness to small perturbations~\cite{barbara2024robust}.
Symmetry regularization leverages the structural symmetry of the robot to constrain the exploration space of the policy, guiding the learning of the optimal policy with symmetry, and improves the coordination and robustness of the robot's motion~\cite{su2024leveraging}.
To address the dynamic mismatch between simulation and reality, recent studies integrate residual action models trained on real-world deployment data~\cite{he2025asap}.
These models can compensate for unmodeled dynamic variations, improve simulation fidelity, and further fine-tune motion policies to improve sim-to-real transfer performance.

Although these methods have achieved remarkable robustness, DR cannot specifically perturb policy vulnerabilities~\cite{tang2020learning, long2024learning}.
Regularization constraints require a trade-off between policy exploration and robust constraints~\cite{shi2024rethinking}.
Residual models are dependent on real-world data, resulting in low implementation efficiency~\cite{salvato2021crossing}.
Furthermore, the approach of improving robustness by accurately identifying vulnerabilities in the policy network and applying targeted perturbations remains underexplored~\cite{chen2024learning}, especially in humanoid robots, where high-dimensional observations and degrees of freedom complicate motion policies.

In this work, we propose a novel \textbf{S}elective \textbf{A}dversarial \textbf{A}ttack for \textbf{R}obust \textbf{T}raining (\textbf{SA2RT}) to enhance humanoid motion policy robustness.
A learnable adversary network identifies policy vulnerabilities and generates targeted perturbations to destabilize the robot with a minimal attack budget.
Through alternating optimization of attack and motion policies, our method continually strengthens the motion policy against
the strongest discovered attacks. 
We conduct extensive experiments on the Unitree G1 humanoid robot~\cite{example_website} on various challenging terrains and agile trajectory tracking.
The results demonstrate that the SA2RT significantly improves the motion performance of humanoid robots in real-world environments.

In summary, the contributions of this work are as follows:
\begin{itemize}
\item A novel selective adversarial attack for robust training is proposed for humanoid robots.
By introducing adversarial attack policies to identify vulnerability of motion policies, effective adversarial samples are generated to enhance the robustness of motion policies.
\item The Selective Attack Policy (SAP) constrained by an attack budget, enhances the vulnerability mitigation of motion policy without inducing conservative degradation.
\item Extensive experimental results on real robots demonstrate that our method significantly improves the long horizon mobility and tracking performance.
\end{itemize}

\begin{figure*}[t]
\centerline{\includegraphics[width=17cm]{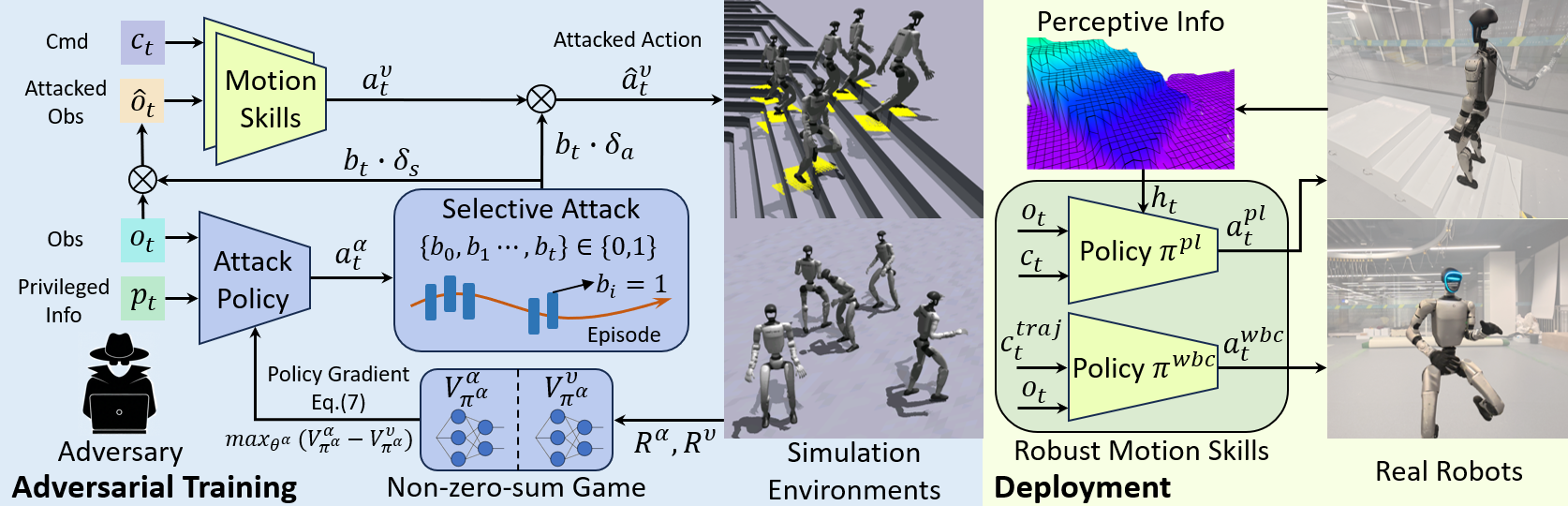}}
\caption{Overview of the SA2RT.
The SAP identifies vulnerabilities in motion states and generates adversarial samples by applying perturbations in both state spaces and action spaces.
Through alternating adversarial training under non-zero-sum game, the motion policy continuously addresses its own vulnerabilities using adversarial samples, enhancing its robustness against perturbations.
During deployment, the robust motion skills are deployed to real robots without requiring the SAP, enabling robust whole-body motion control for humanoid robots.}
\label{fig2}
\end{figure*}

\section{Related Works}\label{Related}
\subsection{Learning-Based Humanoid Motion Control}\label{Related-1}
In recent years, RL-based approaches have achieved impressive results in humanoid robot motion control~\cite{ren2025vb,wei2023learning,radosavovic2024humanoid,cui2024adapting,gu2024advancing}, especially in complex terrain traversal~\cite{zhuang2024humanoid,long2024learning} and whole-body trajectory imitation~\cite{cheng2024expressive,he2024learning}.
Radosavovic et al.~\cite{radosavovic2024humanoid} use causal transformers trained via autoregressive modeling of observations and actions on diverse motion datasets, achieving stable long-distance walking in outdoor environments.
Sun et al.~\cite{sun2025learning} extract terrain features around the robot based on visual perception information to enhance the robot's locomotion capabilities in challenging unstructured terrains~\cite{zhuang2024humanoid}.
Furthermore, by remapping human motion trajectories from teleoperation and motion capture (MoCap) to humanoid robots~\cite{ji2024exbody2,he2024hover}, whole-body imitation learning frameworks have been developed to replicate expressive human-like movements~\cite{ze2025twist}.
He et al.~\cite{he2025asap} compensate for dynamic mismatches by collecting data in real robots to train a residual action model and incorporate it into simulation to fine-tune the policy to align with real-world dynamics.
However, the inherent sensitivity of neural control policies to sensor/actuator noise significantly impairs the stability of humanoid robots during high-dynamic motions~\cite{salvato2021crossing}, failing to maintain long horizon robust mobility and tracking performance.

\subsection{Robust Reinforcement Learning}\label{Related-2}
Robust adversarial training introduces adversarial attacks during training~\cite{moos2022robust}, allowing the agent to learn under uncertain perturbations and improve its robustness in real environments~\cite{huang2017adversarial}.
Adversarial attacks can impose perturbations in the observation space~\cite{zhang2021robust}, action space~\cite{tessler2019action}, reward space~\cite{wang2020reinforcement}, and environmental state space~\cite{kuang2022learning}, effectively simulating various uncertainties that the task policy may encounter in deployment scenarios.
Shi et al.~\cite{shi2024rethinking} use adversarial attacks to evaluate the robustness of RL-based locomotion policies for quadrupedal robots.
\cite{shi2025adversarial} decouples the upper-body and lower-body control of humanoid robots as two independent agents, proposing an adversarial motion and imitation learning framework.
On the other hand, considering the high sensitivity of neural networks to small input perturbations~\cite{zhang2024robust}, some studies have investigated the use of Lipschitz-constrained policy networks~\cite{mysore2021regularizing} to improve robustness against input noise and small perturbations~\cite{kobayashi2022l2c2}.
By regularizing or employing specialized network architectures to constrain the Lipschitz constant of neural networks~\cite{cho2024stability}, the output actions of the policy can be effectively smoothed~\cite{barbara2024robust}, thereby improving the robustness of the policy to perturbations~\cite{chen2024learning}.

\section{Preliminaries}\label{Preliminaries}
\subsection{Reinforcement Learning}\label{Preliminaries-1}
The RL-based control problem is typically formulated as a Markov Decision Process (MDP), where an agent $\pi$ dynamically interacts with the environment through trial-and-error mechanisms to maximize the task-specific objective function.
The MDP is defined as $\mathcal{M}=(\mathcal{S},\mathcal{A},\mathcal{P},R,\gamma)$, where $\mathcal{S}$ is the state space, $\mathcal{A}$ is the action space, $\mathcal{P}:\mathcal{S}\times\mathcal{A}\times\mathcal{S}\rightarrow\mathbb{R}$ is the transition probability, $R(s,a)$ is the reward function, and $\gamma\in[0,1)$ is the discount factor.
At each timestep $t$, the agent $\pi$ receives an observation $s_{t}$ and chooses an action $a_{t}\in\pi(s_{t})$ to obtain the reward $R(s_{t},a_{t})$.
The agent's goal is to maximize the expected cumulative return:
\begin{equation}\label{eq1}
J(\pi)=\mathbb{E}_{\pi}\left[\sum_{t=0}^{\infty}\gamma^{t}R(s_{t},a_{t})\right].
\end{equation}

\subsection{Two-Player Markov Games}\label{Preliminaries-2}
A two-player Markov game~\cite{guo2021adversarial} can be formally represented as ($\mathcal{N},\mathcal{S},\{\mathcal{A}^{i}\}_{i\in\mathcal{N}},\mathcal{T},\{R^{i}\}_{i\in\mathcal{N}},\gamma$), where $\mathcal{N}=\{\pi^{\alpha},\pi^{\upsilon}\}$ represents the set of agents with adversary $\pi^{\alpha}$ and victim $\pi^{\upsilon}$. $\mathcal{S}$ represents the state space of the global environment observations.
$\mathcal{A}^{i}$ represents the action space of the agent $i$. $\mathcal{T}:\mathcal{S}\times\mathcal{A}^{\alpha}\times\mathcal{A}^{\upsilon}\rightarrow\mathbb{R}$ represents the transition probability for any joint actions of both agents. $R^{i}:\mathcal{S}\times\mathcal{A}\rightarrow\mathbb{R}$ is the reward function for the agent $i$.
$\gamma$ is the discount factor.
The state-value function for each agent is defined as a function of the joint policy $\pi(a\vert s)=\prod_{\pi^{i}\in\mathcal{N}}\pi^{i}(a^{i}\vert s)$
\begin{equation}\label{eq2}
V_{\pi}^{i}(s)=\mathbb{E}_{a\sim\pi(a\vert s)}\left[R^{i}(s,a)+\gamma\mathbb{E}_{s'\sim\mathcal{T}}[V_{\pi}^{i}(s')]\right].
\end{equation}
In this Markov game, the victim $\pi^{\upsilon}$ improves its defense capability by maximizing its state value function $V_{\pi}^{\upsilon}(s)$, while the adversary $\pi^{\alpha}$ enhances its attack effect by maximizing its state value function $V_{\pi}^{\alpha}$, thereby enhancing the adversarial robustness.

\section{Methods}\label{Methods}
In this section, we present the SA2RT in detail, which aims to enhance the robustness of humanoid motion policies via selective adversarial attacks.
As shown in Fig.~\ref{fig2}, the framework integrates two learnable modules: motion policy and attack policy.
The motion policy optimizes task performance under perturbations from the attack policy, improving perturbation resistance.
Conversely, the attack policy identifies vulnerable states and selectively applies adversarial perturbations to induce falls.
The two modules are alternately optimized in a game-theoretic manner.

\subsection{Selective Adversarial Attacks}
When deploying RL-based neural controllers, observations are susceptible to natural noise and outliers, causing state estimation errors that lead to suboptimal actions.
Additionally, actuator latency and mechanical wear introduce execution deviations, further degrading motion performance.
To holistically assess policy vulnerabilities from these observation-actuator error cascades, we define the adversarial attack policy’s action space as the joint space of the motion policy’s state and action spaces.
Let $\mathcal{B}_{s}=\{s+\delta_{s}\vert\delta_{s}\in\mathcal{D}_{s}\}$ and $\mathcal{B}_{a}=\{a+\delta_{a}\vert\delta_{a}\in\mathcal{D}_{a}\}$ be bounded perturbation sets for states and actions, respectively, where $\Vert\delta_{s}\Vert\leq\epsilon_{s}$, $\Vert\delta_{a}\Vert\leq\epsilon_{a}$.
Given a pre-trained motion policy $\pi^{m}$ on MDP $\mathcal{M}=(S,A,P,R,\gamma)$, we construct an adversarial MDP $\hat{\mathcal{M}}=(\mathcal{S},\hat{\mathcal{A}},\hat{\mathcal{P}},\hat{\mathcal{R}},\gamma)$ where $\hat{A}=\mathcal{S}\times\mathcal{A}$ represents the joint state-action attack space, $\hat{\mathcal{P}}(s'\vert s,\hat{a})=\mathbb{E}_{a\sim\pi^{m}(\cdot\vert s+\delta_{s})}[\mathcal{P}(s'\vert s,a+\delta_{a})]$, and $\hat{R}(s,a)$ encodes attack objectives.
The Persistent Attack Policy (PAP) in $\hat{\mathcal{M}}$ maximizes attack efficacy while adhering to perturbation bounds $\mathcal{B}_{s}$, $\mathcal{B}_{a}$.
The temporal homogeneity of the PAP prevents it from distinguishing state-specific vulnerability differences and renders it easily detectable, leading to excessive conservatism in the victim policy.

To further improve the stealth and efficiency of attacks, we propose a Selective Attack Policy (SAP).
By introducing an attack-budget constraint, SAP not only determines how to generate optimal perturbations but also identifies vulnerable states where attacks are most effective.
This approach aims to significantly degrade the performance of the motion policy using the fewest attack steps, thereby maximizing impact while minimizing detectability.
In an episode, the motion policy produces a state-action sequence $\{s_{t},a_{t}\}_{t=0}^{T}$.
The SAP strategically selects a subset of timesteps $\mathcal{T}_{adv}\subset\{1,\cdots,T\}$ (where $\vert\mathcal{T}_{adv}\vert\ll T$) to perturb the robot, rather than attacking all timesteps.
Let $\{\delta_{1},\cdots,\delta_{T}\}$ denote a sequence of perturbations sampled from an arbitrary attack policy $\pi^{adv}$, and $N_{a}$ be the attack budget.
The SAP can be formulated as the following optimization problem:
\begin{equation}\label{eq3}
\begin{split}
\max_{b_{0},\cdots,b_{T},\:\pi^{adv}}&\quad \mathbb{E}[\sum_{t=0}^{T}\hat{R}(s_{t},a_{t})] \\
s.t.& \quad [\delta_{s},\delta_{a}]\sim\pi^{adv}(s_{t}), \\
    & \quad a_{t}\sim\pi(s_{t}+b_{t}\delta_{s,t})+b_{t}\delta_{a,t}, \\
    & \quad s_{t+1}\sim\mathcal{P}(s_{t},a_{t}), \\
    & \quad \sum_{t=0}^{T}b_{t}\leq N_{a}.
\end{split}
\end{equation}
The binary variable $b_{t}$ indicates whether the perturbation $\pi^{adv}(s_{t})$ is added to the state $s_{t}$, and step $t$ is a critical step found by the SAP if $b_{t}=1$. To address the attack-budget constraint in problem (\ref{eq3}), we introduce a Lagrange multiplier $\lambda$ to transform the hard budget constraint into a soft penalty term:
\begin{equation}\label{eq4}
\begin{split}
\max_{b_{0},\cdots,b_{T},\:\pi^{adv}}&\quad \mathbb{E}\left[\sum_{t=0}^{T}\hat{R}(s_{t},a_{t})\right] - \lambda\sum_{t=0}^{T}b_{t}.\\
\end{split}
\end{equation}
The $\lambda$ is a hyperparameter that controls the penalty for each attack.
A higher penalty parameter $\lambda$ corresponds to a lower budget constraint $N_{a}$.
We use RL to train SAP in another MDP $\hat{\mathcal{M}}'=(\mathcal{S},\hat{\mathcal{A}}',\hat{\mathcal{P}}',\hat{R}',\gamma')$, where $\hat{\mathcal{A}}'=\{0,1\}\times\hat{A}$, $\hat{\mathcal{P}'}(s'\vert s,\hat{a}')=\mathbb{E}_{a\sim\pi^{m}(\cdot\vert s+b\delta_{s})}[\mathcal{P}(s'\vert s,a+b\delta_{a})]$, $\hat{R}'=\hat{R}-\lambda b$, and $\gamma'=1$.

\subsection{Non-Zero-Sum Adversarial Training}
When the reward function $R^{\alpha}=-R^{\upsilon}$, adversarial training establishes a zero-sum game between the victim $\pi^{\upsilon}$ and the adversary $\pi^{\alpha}$, where both policy updates follow an iterative competitive process to maximize and minimize the shared value function, respectively.
As demonstrated by Perolat et al.~\cite{perolat2015approximate}, the optimal equilibrium value function ensures the equivalence between the minimax equilibrium and the Nash equilibrium in such games:
\begin{equation}\label{eq5}
{V_{\pi}^{\alpha}}^{*}(s)=\min_{\pi^{\upsilon}}\max_{\pi^{\alpha}}V_{\pi}^{\alpha}(s)=\max_{\pi^{\alpha}}\min_{\pi^{\upsilon}}V_{\pi}^{\alpha}(s).
\end{equation}
However, previous analyses of the minimax formulation in adversarial training shows that the attacker learned under the zero-sum game framework does not guarantee continuous improvement in attack performance. This results in a weak adversary dilemma, which in turn leads to a robust overfitting of the victim.

In this work, we formulate the adversarial training between the motion policy $\pi^{m}$ and the attack policy $\pi^{adv}$ as a non-zero-sum game.
The attack policy maximizes the attack reward $R^{adv}$ associated with the robot's safety failure while negatively affecting the motion policy, which can be formally expressed as:
\begin{equation}\label{eq6}
J(\theta_{adv})=\mathop{maximize}\limits_{\theta_{adv}}\:(V_{\pi}^{adv}(s)-V_{\pi}^{m}(s)),
\end{equation}
where $\theta_{adv}$ is the parameter of the attack policy network, $V_{\pi}^{adv}(s)$ and $V_{\pi}^{m}(s)$ represent the attacker's value function and the motion agent's value function, respectively.
A trivial solution to solving Eq.~\eqref{eq6} involves applying policy gradient methods.
However, this solution does not guarantee monotonic changes in both value functions. 
Taking advantage of the work ~\cite{guo2021adversarial}, a new optimization objective that can guarantee the monotonicity of Eq.~\eqref{eq6} is obtained by approximating the value function:
\begin{equation}\label{eq7}
\begin{split}
\mathop{argmax}\limits_{\theta_{adv}}&\:\mathbb{E}_{(a_{t}^{adv},s_{t})\sim\pi^{adv}_{k}}[min(clip(\rho_{t},1-\epsilon,1+\epsilon)A_{t}^{adv},\\
\rho_{t}A_{t}^{adv})&-min(clip(\rho_{t},1-\epsilon,1+\epsilon)A_{t}^{m},\rho_{t}A_{t}^{m})], \\
&\rho_{t}=\frac{\pi^{adv}(a_{t}^{adv}\vert s_{t})}{\pi^{adv}_{k}(a_{t}^{adv}\vert s_{t})}, A_{t}^{adv}=A_{\pi^{adv}_{k}}^{adv}(a_{t}^{adv},s_{t}), \\
&A_{t}^{m}=A_{\pi^{adv}_{k}}^{m}(a_{t}^{adv},s_{t}),
\end{split}
\end{equation}
where $\pi^{adv}_{k}$ denotes the old attack policy and $\epsilon$ represents the clipping parameter in the PPO algorithm~\cite{schulman2017proximal}.
Subsequently, we employ temporal difference learning with two separate neural networks to approximate the value functions of the attack policy and the motion policy.
The policy gradient method is then applied to optimize Eq.~\eqref{eq7}, learning a powerful attack policy.

\begin{algorithm}[t]
\caption{Alternating Adversarial Training}
\label{Algorithm1}
\textbf{Input:} The motion policy $\pi^{m}$ with pre-trained parameters $\theta_{m}$, the attack policy $\pi^{adv}$ with random parameters $\theta_{adv}$, environment $\mathcal{E}$.
\begin{algorithmic}[1]
\FOR{$i=1$ to $N_{iter}$}
\FOR{$j=1$ to $N_{adv}$}
\STATE \{($s_{t}$,$a_{t}^{m}$,$a_{t}^{adv}$,$R_{t}^{m}$,$R_{t}^{adv}$)\}$\leftarrow$ rollout($\mathcal{E}$,$\pi^{m}$,$\pi^{adv}$)
\STATE Update $\theta_{adv}$ with $\mathcal{D}_{adv}:=$\{($s_{t}$,$a_{t}^{adv}$,$R_{t}^{m}$,$R_{t}^{adv}$)\}
\ENDFOR
\FOR{$j=1$ to $N_{m}$}
\STATE \{($s_{t}$,$a_{t}^{m}$,$a_{t}^{adv}$,$R_{t}^{m}$,$R_{t}^{adv}$)\}$\leftarrow$ rollout($\mathcal{E}$,$\pi^{m}$,$\pi^{adv}$)
\STATE Update $\theta_{m}$ with $\mathcal{D}_{m}:=$\{($s_{t}$,$a_{t}^{m}$,$R_{t}^{m}$)\}
\ENDFOR
\ENDFOR
\end{algorithmic}
\end{algorithm}

We employ an alternating training procedure to optimize the motion policy and the attack policy.
Algorithm~\ref{Algorithm1} outlines our approach in detail. 
The motion policy is initialized from the pre-trained policy, while the attack policy is randomly initialized.
In $N_{adv}$ iterations, the motion policy is fixed and embedded into the environment, and the attack policy is optimized to find the intrinsic weaknesses of the motion policy and impose attacks. 
Then in $N_{m}$ iterations, the attack policy is fixed, enabling the motion policy to be further optimized for resilience against adversarial attacks. 
Through dynamic alternating optimization over $N_{iter}$ iterations, our framework achieves online robust adversarial training. 
This paradigm promotes the motion policy to adaptively enhance its robustness against various adversarial attacks.

\section{Experiments}
\subsection{Experimental Setup}\label{Experiments-A}
We apply the SA2RT to two humanoid robot tasks, Perceptive Locomotion (PL) and Whole-Body Control (WBC), and implementation details are provided in the Appendix.
Through comparative experiments and ablation studies, we address the following research questions:
(1) Can adversarial attacks induce stronger perturbations than domain randomization?
(2) Can SA2RT improve the robustness of motion policies?
(3) Does SA2RT contribute to the motion performance of real robots?

\subsubsection{Baselines}
We design four different training settings for comparative analysis, and each motion policy is represented using the following notation:
\begin{itemize}
\item \textbf{PL/WBC-CE}: The motion policies trained in Clean Environments (CE), without DR and adversarial attacks.
Specifically, PL-CE represents the baseline of the perceptive locomotion policy, while WBC-CE represents the baseline of the whole-body control policy.
\item \textbf{PL/WBC-DR}: Applying DR in training environments.
\item \textbf{PL/WBC-PAP}: Robust adversarial training with persistent attack policies.
\item \textbf{PL/WBC-SAP}: Robust adversarial training with selective attack policies.
\end{itemize}
 
\subsubsection{Metrics}
To evaluate the performance of motion policies, we use several metrics:
\begin{itemize}
\item The mean linear velocity error $E_{vel}(m/s)$ and the mean angular velocity error $E_{ang}(rad/s)$ are used to evaluate the velocity tracking accuracy.
\item The gravitational projection component $E_{g}(m/s^{2})$ is used to evaluate the dynamic stability of the robot.
\item The Mean Per Keybody Position Error (MPKPE) $E_{mpkpe}^{upper}(m)$ and $E_{mpkpe}^{lower}(m)$ are used to evaluate the keypoint tracking accuracy.
\item The Mean Per Joint Position Error (MPJPE) $E_{mpjpe}^{upper}(rad)$ and $E_{mpjpe}^{lower}(rad)$ are used to evaluate the joint position tracking ability.
\item The success rate $R_{sr}(\%)$ for robot survival in complex terrain traversal and whole-body trajectory tracking.
\end{itemize}

\subsection{Simulation Results}\label{Experiments-B}
\subsubsection{Effectiveness of Adversarial Perturbations}
To ensure the comparability of experimental results, we maintain consistent perturbation spaces and thresholds for both DR and SAP during training.
We then evaluate the task success rates of motion policies under different training settings by separately applying DR and SAP.
As shown in Table~\ref{table1}, DR sharply reduces PL/WBC-CE’s success rate yet leaves adversarially trained PL/WBC-SAP unaffected.
Conversely, the SAP compromises both PL/WBC-CE and PL/WBC-DR.
The results show that simple DR cannot ensure the robustness of the motion policy, while SAP effectively exposes vulnerabilities and introduces stronger perturbations.

\begin{table}[t]
\centering
\begin{tabular*}{0.46\textwidth}{w{c}{0.03\textwidth}@{\extracolsep{\fill}}c@{\extracolsep{\fill}}c@{\extracolsep{\fill}}c@{}}
\toprule
\multirow{2}{*}{Envs} & \multicolumn{3}{c}{Motion Policies $R_{sr}(\%)$} \\
\cline{2-4}
\addlinespace[0.5ex] 
    & PL/WBC-CE & PL/WBC-DR & PL/WBC-SAP\\
\midrule
DR & $77.3\pm7.3$ & $94.4\pm5.7$ & $97.4\pm2.1$ \\
SAP & $\mathbf{0.0\pm0.0}$ & $\mathbf{0.0\pm0.0}$ & $\mathbf{95.7\pm6.4}$ \\
\bottomrule
\end{tabular*}
\caption{Success Rates in DR and SAP Environments.}
\label{table1}
\end{table}

\begin{table}[t]
\centering
\begin{tabular*}{0.46\textwidth}{w{c}{0.03\textwidth}w{c}{0.045\textwidth}@{\hspace{10pt}}@{\extracolsep{\fill}}c@{\extracolsep{\fill}}c@{\extracolsep{\fill}}c@{}}
\toprule
\multirow{2}{*}{Envs} & \multirow{2}{*}{Policies} & \multicolumn{3}{c}{Metrics} \\
\cline{3-5}
\addlinespace[0.2ex] 
 &   & $E_{vel}(m/s)$ & $E_{ang}(rad/s)$ & $E_{g}(m/s^{2})$ \\
\midrule
\multirow{2}{*}{CE} & PL-DR  &                                          $0.25\pm0.03$ &                                     $0.29\pm0.02$ &                                     $0.34\pm0.01$ \\
                    & PL-SAP & $\mathbf{0.19\pm0.02}$ & $\mathbf{0.22\pm0.03}$ & $\mathbf{0.30\pm0.01}$ \\
\hline
\multirow{2}{*}{DR} & PL-DR  & 
                    $0.28\pm0.02$ & 
                    $0.34\pm0.03$ & 
                    $0.42\pm0.01$ \\
                    & PL-SAP & $\mathbf{0.21\pm0.03}$ & $\mathbf{0.27\pm0.02}$ & $\mathbf{0.36\pm0.01}$ \\
\bottomrule
\end{tabular*}
\caption{Robustness Analysis of PL Policies.}
\label{table2}
\end{table}

\subsubsection{Robustness of Motion Policies}
\begin{figure}[t]
\centerline{\includegraphics[width=8.2cm]{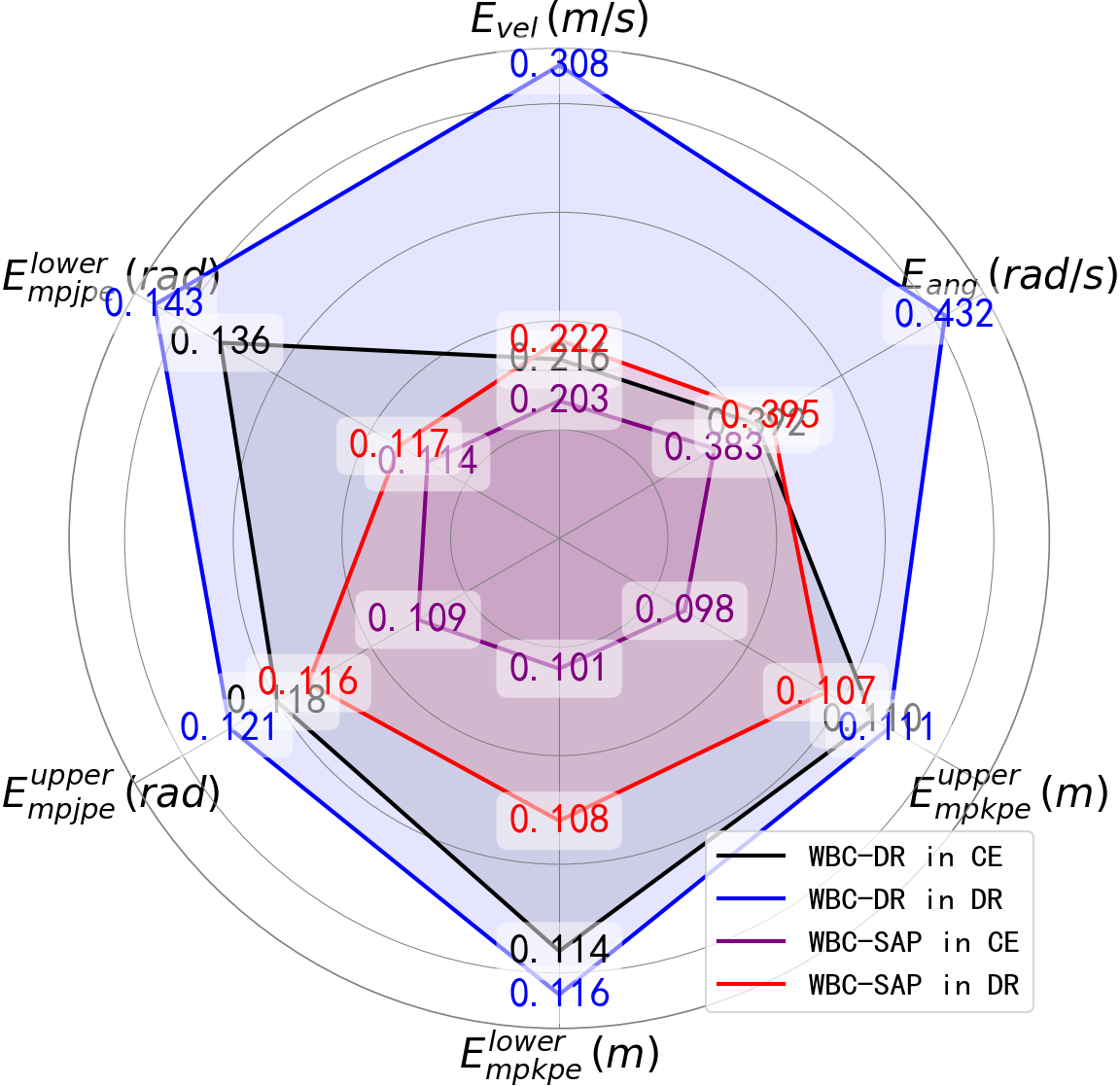}}
\caption{Performance analysis of whole-body control.
Trajectory tracking errors of WBC-DR and WBC-SAP are evaluated in clean environments and DR environments.
WBC-SAP outperforms WBC-DR across all evaluation metrics, demonstrating that the SA2RT effectively enhances the robustness and tracking performance of WBC policies.
}
\label{fig3}
\end{figure}

We evaluate motion policies trained with different methods in both clean and DR environments.
For the perceptive locomotion task, we evaluate velocity tracking and gravitational projection component on flat ground, slopes, stairs, and discrete terrains.
As shown in Table~\ref{table2}, PL-SAP outperforms PL-DR, 
indicating that that the SA2RT can effectively address vulnerabilities in motion policies, enhancing robustness and task performance.
For whole-body control, we evaluate trajectory tracking performance, as shown in Fig.~\ref{fig3}.
WBC-SAP outperforms WBC-DR across all evaluation metrics, particularly in velocity tracking.
Experimental results demonstrate that the SA2RT effectively enhances the robot’s trajectory imitation performance, enabling it to maintain body stability in perturbed environments without compromising tracking accuracy.

\begin{figure}[!ht]
\centering  
\subfigure[]{
\label{fig4-1}
\includegraphics[width=4.0cm]{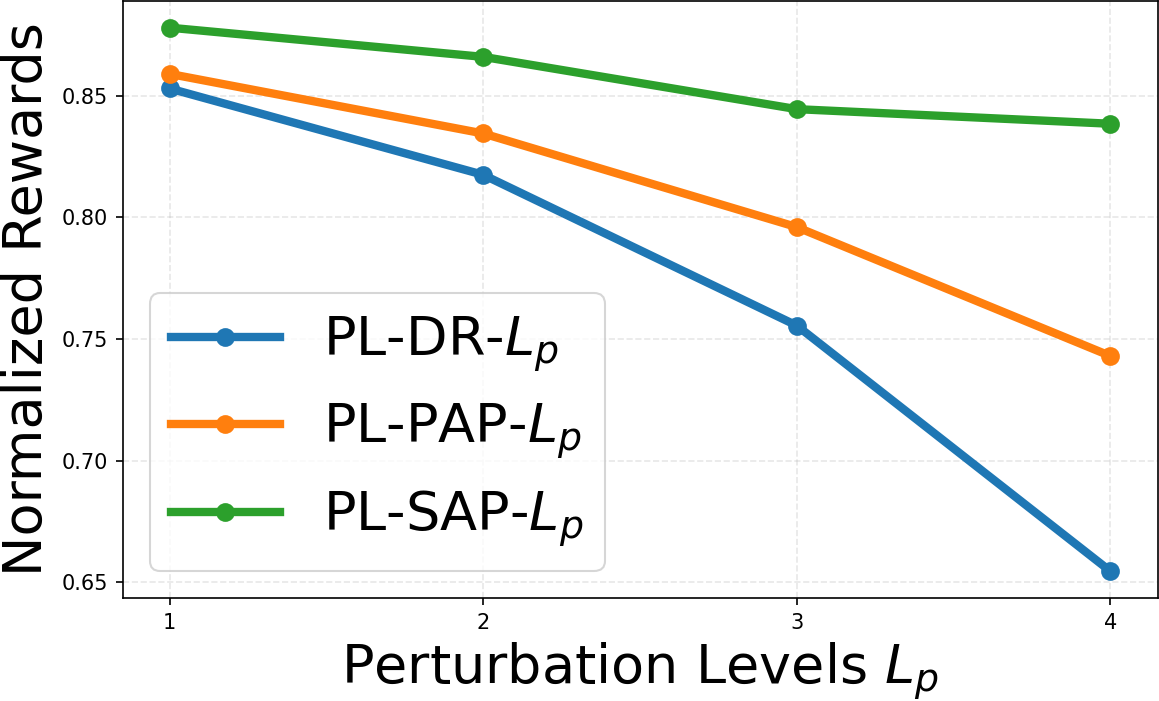}}
\subfigure[]{
\label{fig4-2}
\includegraphics[width=4.0cm]{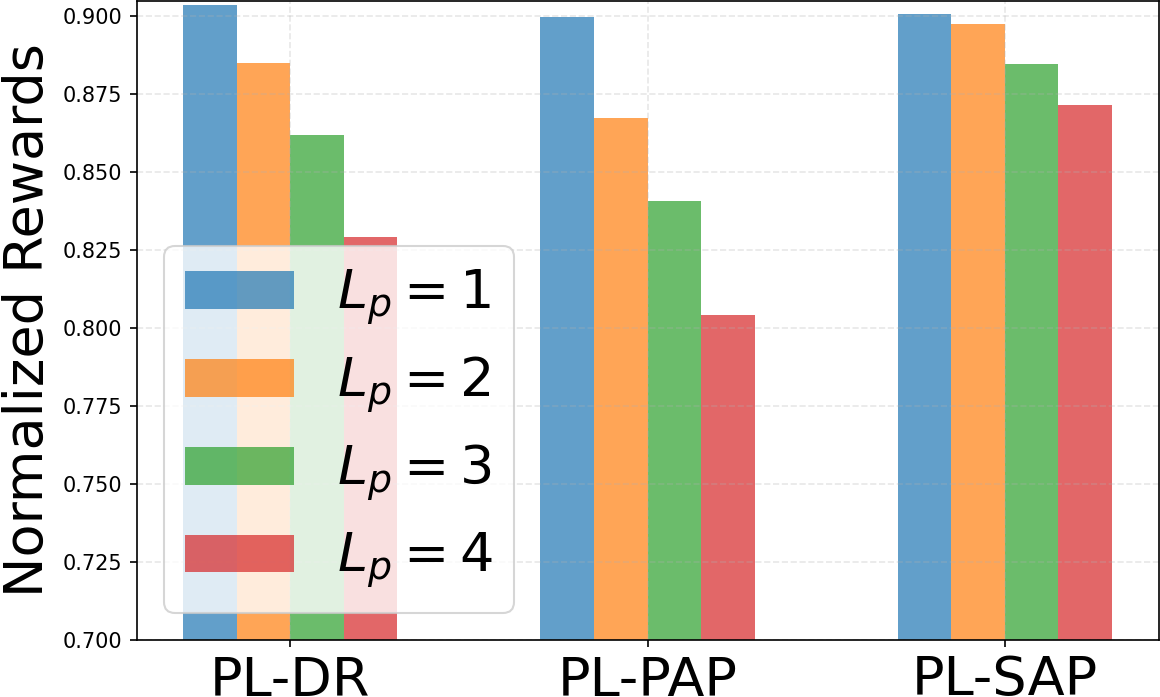}}
\caption{Impact of different attack policies on motion policy performance.
(a) Rewards of motion policies learned via different attack policies under varying perturbation levels $L_{p}$.
(b) Performance comparison in unperturbed environments for motion policies trained under different $L_{p}$.
}
\label{fig4}
\end{figure}

\subsubsection{Ablation Study}
To evaluate the impact of SAP on the performance of motion policies, we compare the performance of motion policies adversarially trained by PAP and SAP under different perturbation levels $L_{p}\in\{1,2,3,4\}$.
For each level, we train PL-PAP-$L_{p}$ and PL-SAP-$L_{p}$ policies and record their normalized rewards, as shown in Fig.~\ref{fig4-1}.
The results demonstrate that adversarial training outperforms DR in robustness.
As the perturbation level $L_{p}$ increases, normalized rewards of PL-SAP-$L_{p}$ decrease less than those of PL-PAP-$L_{p}$, indicating that SAP's selective attacks on vulnerable states prevent the excessive conservatism from invalid stable-state attacks.
In addition, we compare the performance of adversarially trained motion policies under different perturbation levels $L_{P}$, as shown in Fig.~\ref{fig4-2}.
PL-SAP-$L_{p}$ exhibits slower performance degradation compared to PL-PAP-$L_{p}$, while the degradation of PL-PAP-$L_{p}$ exceeds that of PL-DR.
The results indicate that while PAP exhibits stronger attack capabilities, its full-time attack induces a significant shift in the state distribution of the motion policy, leading to a marked decline in the policy's performance in low-perturbation environments.
In contrast, SAP only attacks the vulnerable states of the motion policy, featuring certain high efficiency and stealthiness, and effectively balances the motion policy's performance and robustness.

\begin{figure}[t]
\centerline{\includegraphics[width=8.2cm]{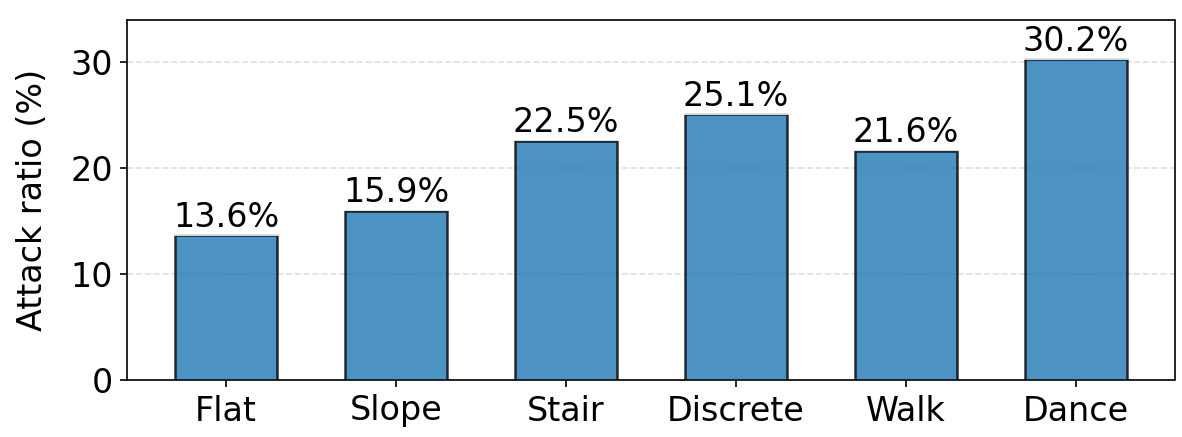}}
\caption{SAP's attack ratio varies significantly across motion tasks.
As the difficulty of the task increases, the attack rate gradually increases.
}
\label{fig5}
\end{figure}

\begin{figure}[!ht]
\centering  
\subfigure[]{
\label{fig6-1}
\includegraphics[width=3.9cm]{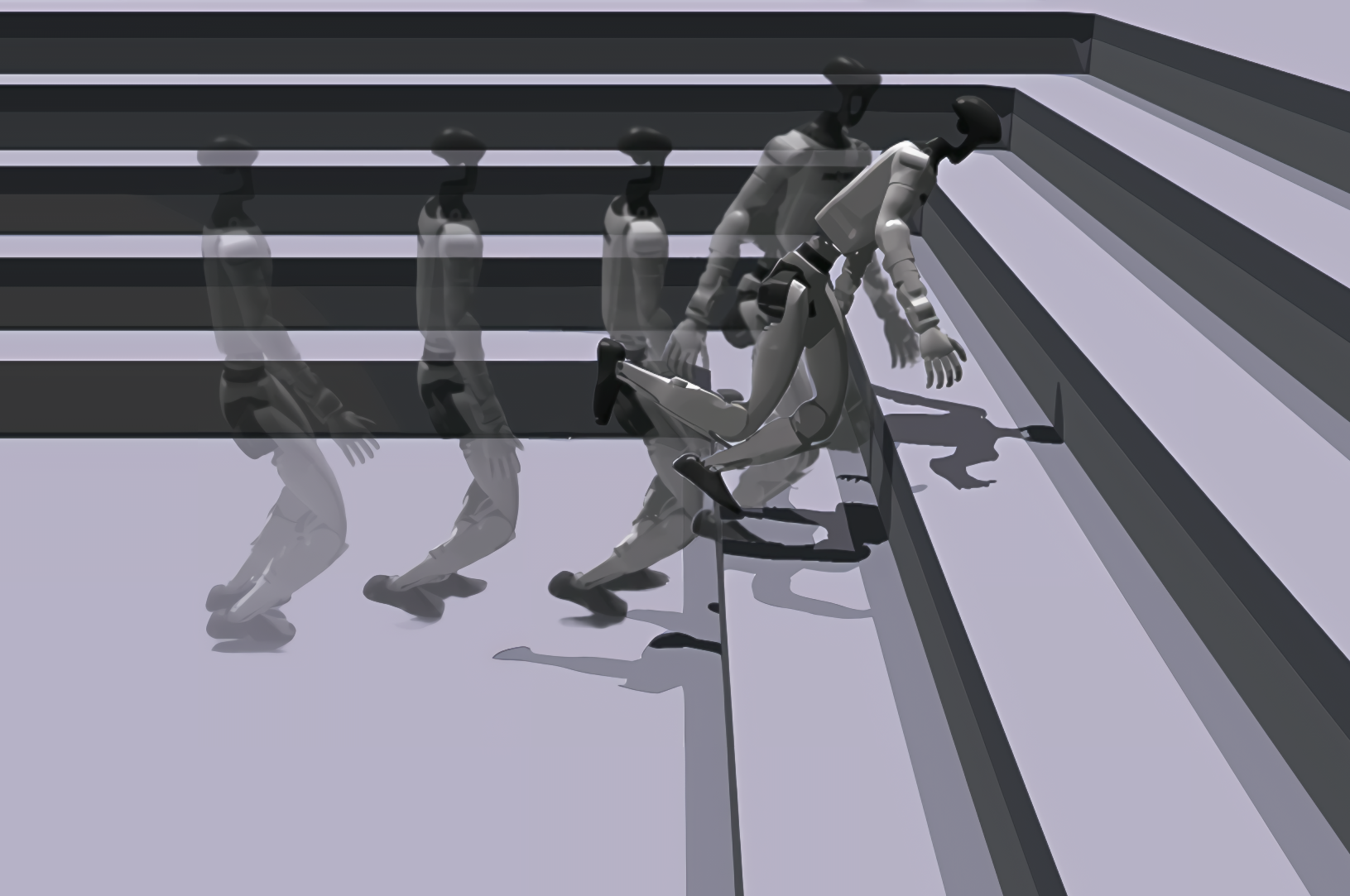}}
\subfigure[]{
\label{fig6-2}
\includegraphics[width=4.2cm]{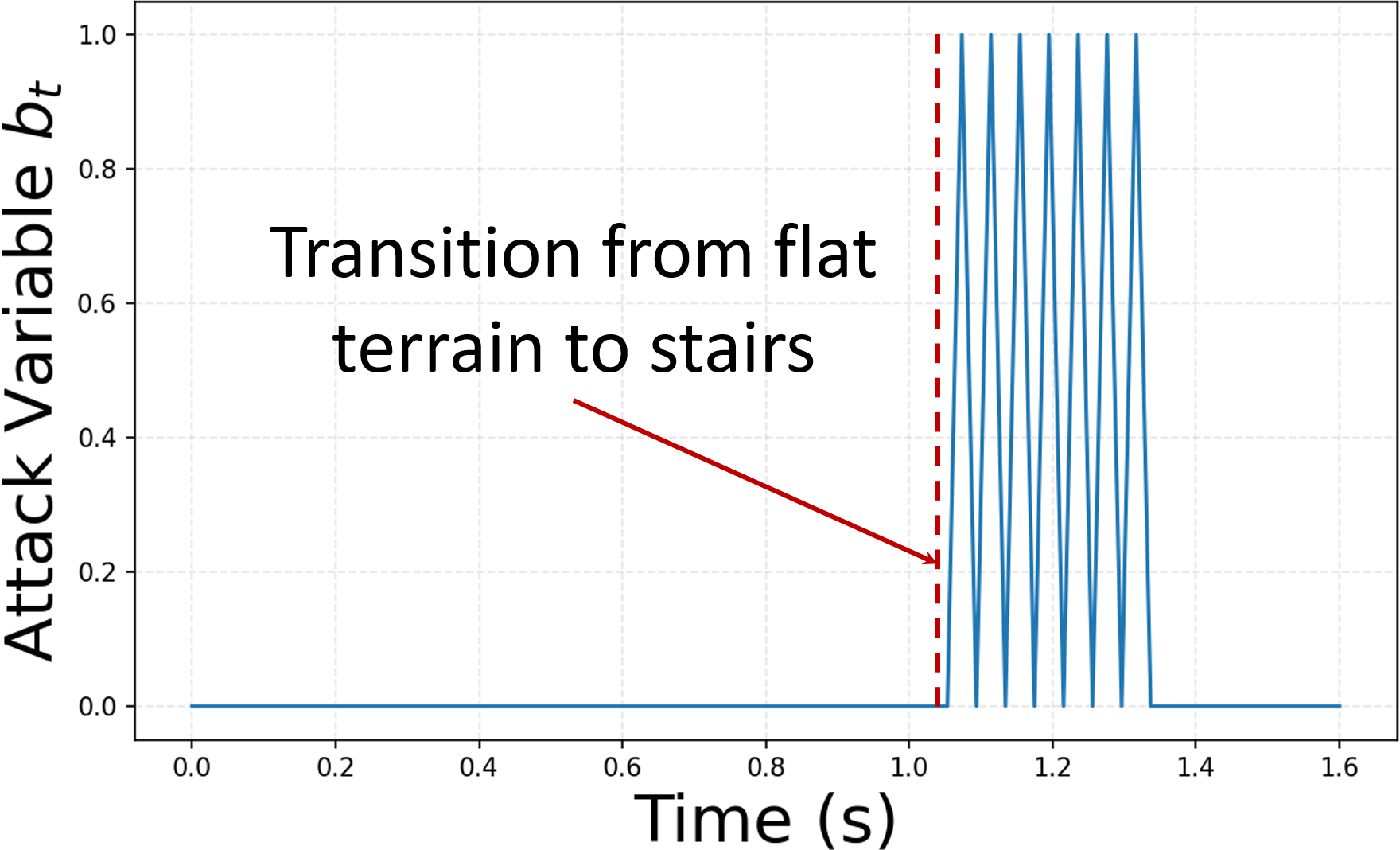}}
\caption{SAP attack sequences during the robot's transition from flat terrain to stairs.
(a) Robot motion states visualization.
(b) Attack variable sequence.
As the robot readies to traverse stairs, SAP identifies the state's vulnerability, applies a few attacks, and successfully induces a fall.}
\label{fig6}
\end{figure}

\begin{figure}[!ht]
\centerline{\includegraphics[width=8.2cm]{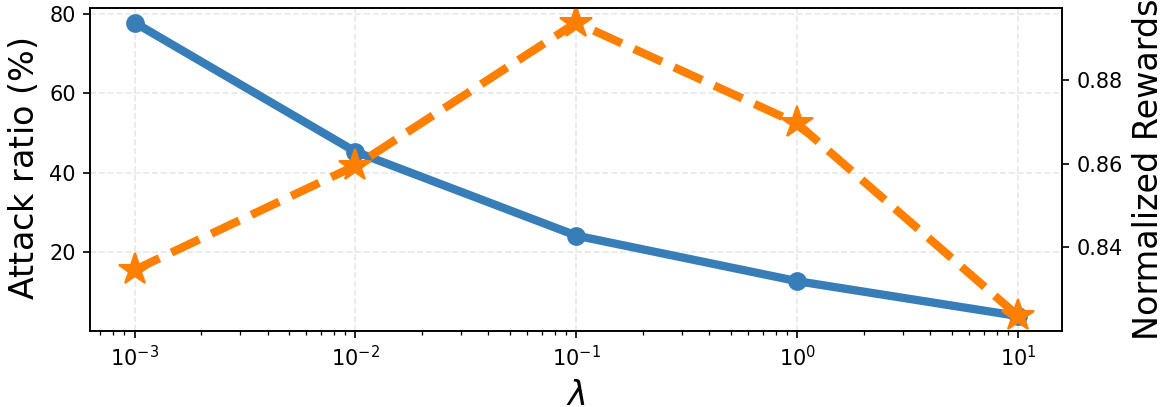}}
\caption{
Comparison of WBC-SAP's attack ratio and performance across parameter $\lambda$.
The blue line represents the attack ratio and the orange line represents the reward.
As $\lambda$ increases, the attack ratio decreases, but the performance shows a trend of first increasing and then decreasing.
}
\label{fig7}
\end{figure}

\subsubsection{Analysis of Selective Attack Policy}
We evaluate SAP's attack ratio across different motion tasks (Fig.~\ref{fig5}), testing perceptive locomotion on flat ground, slopes, stairs, and discrete terrain, while categorizing whole-body control into simple walking and dynamic dancing trajectories.
The results demonstrate that the attack ratio increases with task difficulty, confirming SAP's capability to learn and execute targeted attacks.
In addition, we analyze SAP's attack behavior during the robot's transition from flat terrain to stairs (Fig.~\ref{fig6}).
The SAP selectively attacks vulnerable transition phases, thereby inducing falls.
This indicates that SAP can clearly distinguish differences in vulnerability across the robot's various motion states and apply targeted perturbations, prompting the motion policy to continuously compensate for vulnerable states during adversarial training and enhancing its robustness.

\subsubsection{Analyzing Hyperparameter $\lambda$}
In selective adversarial training, hyperparameter $\lambda$ regulates the number of adversarial attacks.
We analyze WBC-SAP's attack ratio and motion performance under different $\lambda$.
As illustrated in Fig.~\ref{fig7}, as $\lambda$ increases, the attack ratio drops significantly, while motion performance first rises then falls.
When $\lambda$ increases from $10^{-3}$ to $10^{-1}$, SAP reduces attacks while identifying robot motion vulnerabilities, improving attack efficiency and policy robustness.
However, further increasing $\lambda$ from $10^{-1}$ to $10^{1}$, leads to excessive penalties on the number of attacks, thereby severely reducing the attack ratio.
This weakens targeted attacks and hinders improvements in robustness.
Thus, $\lambda$ can adjust the number of attacks, constraining the SAP to learn targeted attacks and avoiding inefficient persistent attacks that compromise task performance.

\subsection{Real-World Experiments}\label{Experiments-C}
\subsubsection{Complex Terrain Traversability}
\begin{figure}[!ht]
\centering  
\subfigure[]{
\label{fig8-1}
\includegraphics[width=4.0cm]{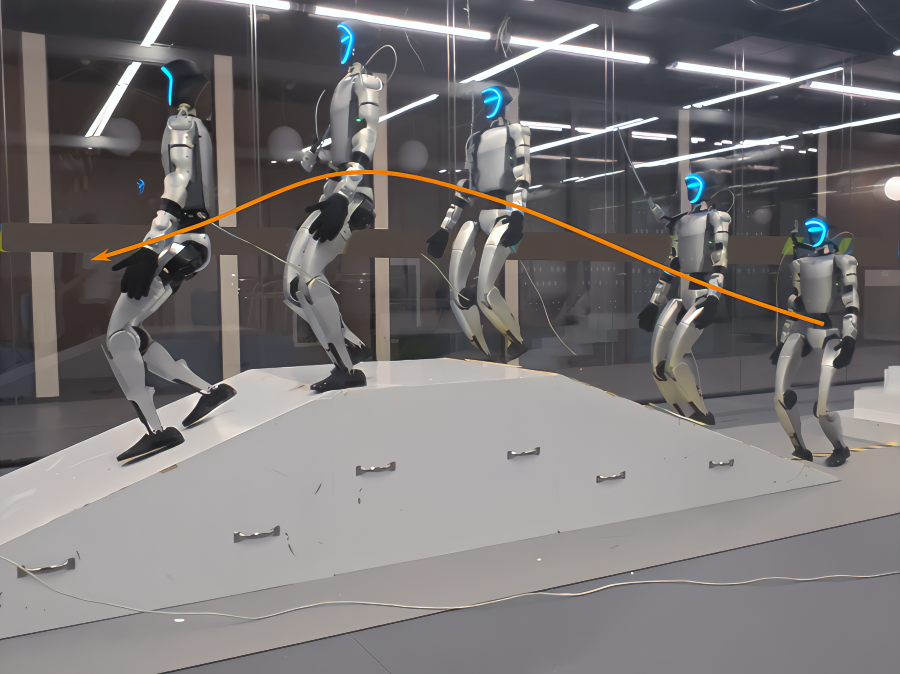}}
\subfigure[]{
\label{fig8-2}
\includegraphics[width=4.0cm]{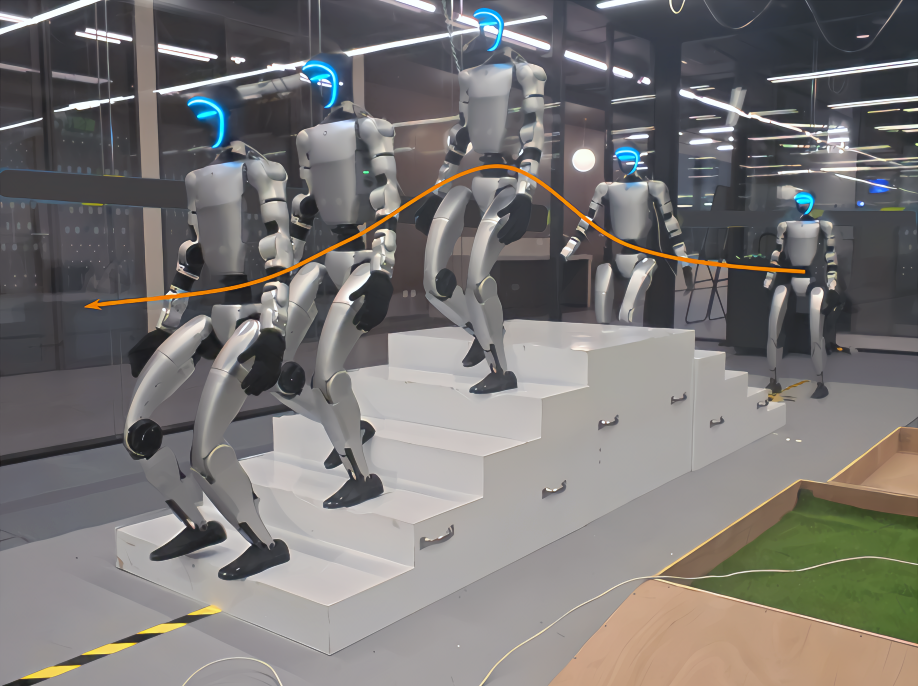}}
\subfigure[]{
\label{fig8-3}
\includegraphics[width=8.2cm]{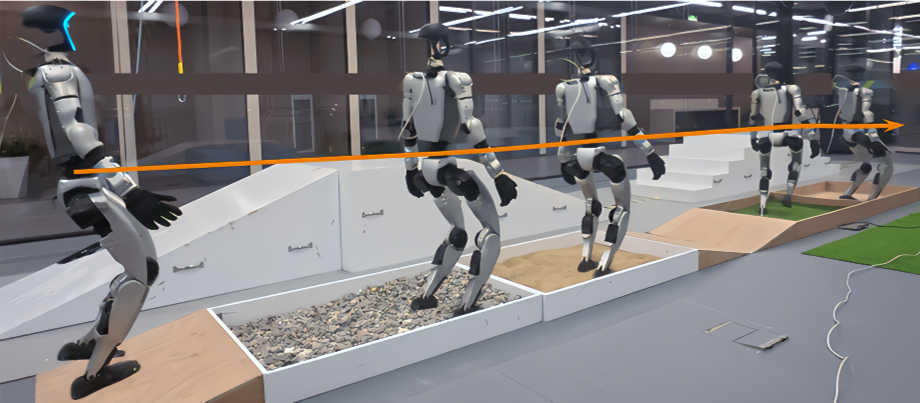}}
\caption{Evaluating the motion performance of PL-SAP in traversing complex terrain.
(a) The robot traverses $22\,^{\circ}$ slopes.
(b) The robot traverses $16\,cm$ stairs.
(c) The robot traverses gravel and sandy terrain.
PL-SAP exhibits higher success rates in complex terrain traversal, indicating the SA2RT effectively enhances motion policy robustness.}
\label{fig8}
\end{figure}

To evaluate the terrain traversability of PL-SAP, we construct a standardized test environment with challenging terrains (stairs, slopes, gravel, sand, grass) and compare traversal success rates between PL-DR and PL-SAP, as illustrated in Table~\ref{table3}.
Each terrain is tested in $5$ groups, with $10$ trials per group.
Benefiting from effective vulnerability identification and targeted perturbation attacks, the SA2RT improves the traversal success rates of the real robot by $40\%$.
Fig.~\ref{fig8} shows snapshots of complex terrain traversal.
PL-SAP exhibit significantly fewer foot trips and slips than PL-DR on stairs and slopes.
Additionally, PL-SAP effectively adapts to thin, sparse ridges in terrain connections (hard to perceive via elevation maps), avoiding trips and maintaining stability, whereas PL-DR often falls here.
These results demonstrate that the SA2RT significantly enhances the robustness of motion policies.

\begin{table}[t]
\centering
\begin{tabular*}{0.46\textwidth}{w{c}{0.06\textwidth}@{\extracolsep{\fill}}c@{\extracolsep{\fill}}c@{\extracolsep{\fill}}c@{\extracolsep{\fill}}c@{\extracolsep{\fill}}c@{}}
\toprule
\multirow{2}{*}{Policies} & \multicolumn{5}{c}{Terrains Traversal $R_{sr}(\%)$} \\
\cline{2-6}
\addlinespace[0.2ex] 
 & Slopes & Stairs & Gravel & Sands & Grass \\
\midrule
PL-DR & $54\pm6.9$ & $47\pm7.5$ & $71\pm4.3$ & $73\pm4.5$ & $82\pm4.9$\\
PL-SAP & $\mathbf{90\pm3.7}$ & $\mathbf{84\pm5.4}$ & $\mathbf{93\pm2.1}$ & $\mathbf{90\pm2.0}$ & $\mathbf{95\pm1.6}$\\
\bottomrule
\end{tabular*}
\caption{Success Rates of Complex Terrain Traversal.}
\label{table3}
\end{table}

\begin{figure}[!ht]
\centerline{\includegraphics[width=8.2cm]{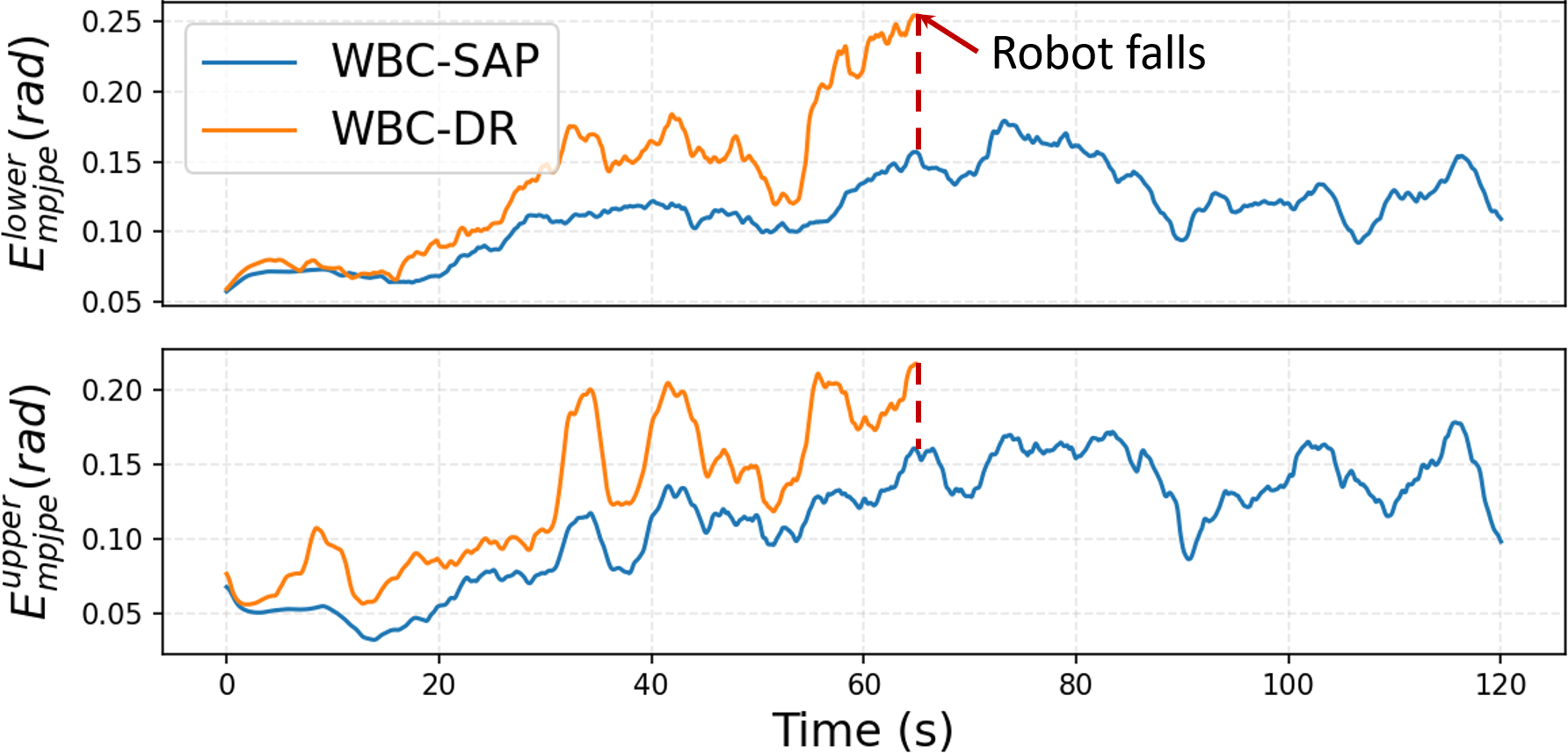}}
\caption{The Mean Per Joint Position Error of each frame in the dance trajectory imitation. Due to the insufficient robustness of PL-DR, the robot loses balance and terminates prematurely at $65$ seconds.
PL-SAP effectively reduces trajectory tracking errors and enables complete long horizon trajectory tracking.}
\label{fig9}
\end{figure}

\subsubsection{Whole-Body Trajectory Imitation Performance}
We further evaluate the performance of the whole-body control on the real robot by tracking $120$ seconds of dynamic dance motion.
The two joint trajectory tracking indicators $E_{mpjpe}^{lower}$ and $E_{mpjpe}^{upper}$ are tested, as shown in Fig.~\ref{fig9}.
Due to the high dynamic balancing demands of the dance trajectory, WBC-DR exhibits progressively accumulating tracking errors, ultimately causing instability and premature termination after only $65$ seconds of execution.
In contrast, WBC-SAP effectively reduces trajectory tracking error by $32\%$ and successfully completes the $120$ seconds motion trajectory.
Snapshots of the imitation of dance motions are presented in Fig.~\ref{fig1}, demonstrating the robot's ability to dynamically regulate the balance of the whole body while performing complex trajectory tracking.
These results indicate that the SA2RT significantly enhances the robustness of motion policies and effectively bridges the sim-to-real gap in dynamic motion control.

\section{Conclusions}
In this paper, we propose a novel selective adversarial attack for robust training that aims to enhance the robustness of RL-based motion skills and improve long horizon mobility and tracking performance in real robots.
By introducing the selective attack policy, which is capable of identifying vulnerable states and imposing targeted perturbations under an attack-budget constraint, the motion policies are strengthened through dynamic non-zero-sum game optimization.
Extensive experiments conducted on the Unitree G1 humanoid robot have verified the superiority of the SA2RT.
Future work will explore potential perturbation attacks in the interaction between the robot and the environment to enhance the robot's adaptability to dynamic environments.
In addition, optimizing the attack-budget constraint to further improve the trade-off between attack effectiveness and strategy agility remains a promising direction.

\section{Acknowledgments}
This work was supported by the National Natural Science Foundation of China (Grant No. 92248303 and No. 62373242), the Shanghai Municipal Science and Technology Major Project (Grant No. 2021SHZDZX0102), and the Fundamental Research Funds for the Central Universities.


\begin{thebibliography}{47}
\providecommand{\natexlab}[1]{#1}

\bibitem[{Barbara, Wang, and Manchester(2024)}]{barbara2024robust}
Barbara, N.~H.; Wang, R.; and Manchester, I.~R. 2024.
\newblock On Robust Reinforcement Learning with Lipschitz-Bounded Policy Networks.
\newblock \emph{arXiv preprint arXiv:2405.11432}.

\bibitem[{Chen et~al.(2024)Chen, He, Wang, Liao, Ze, Li, Sastry, Wu, Sreenath, Gupta et~al.}]{chen2024learning}
Chen, Z.; He, X.; Wang, Y.-J.; Liao, Q.; Ze, Y.; Li, Z.; Sastry, S.~S.; Wu, J.; Sreenath, K.; Gupta, S.; et~al. 2024.
\newblock Learning smooth humanoid locomotion through lipschitz-constrained policies.
\newblock \emph{arXiv preprint arXiv:2410.11825}.

\bibitem[{Cheng et~al.(2024)Cheng, Ji, Chen, Yang, Yang, and Wang}]{cheng2024expressive}
Cheng, X.; Ji, Y.; Chen, J.; Yang, R.; Yang, G.; and Wang, X. 2024.
\newblock Expressive whole-body control for humanoid robots.
\newblock \emph{arXiv preprint arXiv:2402.16796}.

\bibitem[{Cho and Kim(2024)}]{cho2024stability}
Cho, N.; and Kim, Y. 2024.
\newblock On the stability of Lipschitz continuous control problems and its application to reinforcement learning.
\newblock \emph{arXiv preprint arXiv:2404.13316}.

\bibitem[{Cui et~al.(2024)Cui, Li, Huang, Qin, Zhang, Zheng, Tang, Hu, Yan, Chen et~al.}]{cui2024adapting}
Cui, W.; Li, S.; Huang, H.; Qin, B.; Zhang, T.; Zheng, L.; Tang, Z.; Hu, C.; Yan, N.; Chen, J.; et~al. 2024.
\newblock Adapting humanoid locomotion over challenging terrain via two-phase training.
\newblock In \emph{8th Annual Conference on Robot Learning}.

\bibitem[{Fujimoto et~al.(2024)Fujimoto, Suetterlein, Chatterjee, and Ganguly}]{fujimoto2024assessing}
Fujimoto, T.; Suetterlein, J.; Chatterjee, S.; and Ganguly, A. 2024.
\newblock Assessing the Impact of Distribution Shift on Reinforcement Learning Performance.
\newblock \emph{arXiv preprint arXiv:2402.03590}.

\bibitem[{Gu, Wang, and Chen(2024)}]{gu2024humanoid}
Gu, X.; Wang, Y.-J.; and Chen, J. 2024.
\newblock Humanoid-Gym: Reinforcement Learning for Humanoid Robot with Zero-Shot Sim2Real Transfer.
\newblock \emph{arXiv preprint arXiv:2404.05695}.

\bibitem[{Gu et~al.(2024)Gu, Wang, Zhu, Shi, Guo, Liu, and Chen}]{gu2024advancing}
Gu, X.; Wang, Y.-J.; Zhu, X.; Shi, C.; Guo, Y.; Liu, Y.; and Chen, J. 2024.
\newblock Advancing humanoid locomotion: Mastering challenging terrains with denoising world model learning.
\newblock \emph{arXiv preprint arXiv:2408.14472}.

\bibitem[{Gu et~al.(2025)Gu, Li, Shen, Yu, Xie, McCrory, Cheng, Shamsah, Griffin, Liu et~al.}]{gu2025humanoid}
Gu, Z.; Li, J.; Shen, W.; Yu, W.; Xie, Z.; McCrory, S.; Cheng, X.; Shamsah, A.; Griffin, R.; Liu, C.~K.; et~al. 2025.
\newblock Humanoid Locomotion and Manipulation: Current Progress and Challenges in Control, Planning, and Learning.
\newblock \emph{arXiv preprint arXiv:2501.02116}.

\bibitem[{Guo et~al.(2021)Guo, Wu, Huang, and Xing}]{guo2021adversarial}
Guo, W.; Wu, X.; Huang, S.; and Xing, X. 2021.
\newblock Adversarial policy learning in two-player competitive games.
\newblock In \emph{International conference on machine learning}, 3910--3919. PMLR.

\bibitem[{He et~al.(2025)He, Gao, Xiao, Zhang, Wang, Wang, Luo, He, Sobanbab, Pan et~al.}]{he2025asap}
He, T.; Gao, J.; Xiao, W.; Zhang, Y.; Wang, Z.; Wang, J.; Luo, Z.; He, G.; Sobanbab, N.; Pan, C.; et~al. 2025.
\newblock ASAP: Aligning Simulation and Real-World Physics for Learning Agile Humanoid Whole-Body Skills.
\newblock \emph{arXiv preprint arXiv:2502.01143}.

\bibitem[{He et~al.(2024{\natexlab{a}})He, Luo, Xiao, Zhang, Kitani, Liu, and Shi}]{he2024learning}
He, T.; Luo, Z.; Xiao, W.; Zhang, C.; Kitani, K.; Liu, C.; and Shi, G. 2024{\natexlab{a}}.
\newblock Learning human-to-humanoid real-time whole-body teleoperation.
\newblock In \emph{2024 IEEE/RSJ International Conference on Intelligent Robots and Systems (IROS)}, 8944--8951. IEEE.

\bibitem[{He et~al.(2024{\natexlab{b}})He, Xiao, Lin, Luo, Xu, Jiang, Kautz, Liu, Shi, Wang et~al.}]{he2024hover}
He, T.; Xiao, W.; Lin, T.; Luo, Z.; Xu, Z.; Jiang, Z.; Kautz, J.; Liu, C.; Shi, G.; Wang, X.; et~al. 2024{\natexlab{b}}.
\newblock Hover: Versatile neural whole-body controller for humanoid robots.
\newblock \emph{arXiv preprint arXiv:2410.21229}.

\bibitem[{Huang et~al.(2017)Huang, Papernot, Goodfellow, Duan, and Abbeel}]{huang2017adversarial}
Huang, S.; Papernot, N.; Goodfellow, I.; Duan, Y.; and Abbeel, P. 2017.
\newblock Adversarial attacks on neural network policies.
\newblock \emph{arXiv preprint arXiv:1702.02284}.

\bibitem[{Ji et~al.(2024)Ji, Peng, Liu, Li, Yang, Cheng, and Wang}]{ji2024exbody2}
Ji, M.; Peng, X.; Liu, F.; Li, J.; Yang, G.; Cheng, X.; and Wang, X. 2024.
\newblock Exbody2: Advanced expressive humanoid whole-body control.
\newblock \emph{arXiv preprint arXiv:2412.13196}.

\bibitem[{Kobayashi(2022)}]{kobayashi2022l2c2}
Kobayashi, T. 2022.
\newblock L2c2: Locally lipschitz continuous constraint towards stable and smooth reinforcement learning.
\newblock In \emph{2022 IEEE/RSJ International Conference on Intelligent Robots and Systems (IROS)}, 4032--4039. IEEE.

\bibitem[{Kuang et~al.(2022)Kuang, Lu, Wang, Zhou, Li, and Li}]{kuang2022learning}
Kuang, Y.; Lu, M.; Wang, J.; Zhou, Q.; Li, B.; and Li, H. 2022.
\newblock Learning robust policy against disturbance in transition dynamics via state-conservative policy optimization.
\newblock In \emph{Proceedings of the AAAI Conference on Artificial Intelligence}, volume~36, 7247--7254.

\bibitem[{Liang et~al.(2022)Liang, Sun, Zheng, and Huang}]{liang2022efficient}
Liang, Y.; Sun, Y.; Zheng, R.; and Huang, F. 2022.
\newblock Efficient adversarial training without attacking: Worst-case-aware robust reinforcement learning.
\newblock \emph{Advances in Neural Information Processing Systems}, 35: 22547--22561.

\bibitem[{Long et~al.(2024)Long, Ren, Shi, Wang, Huang, Luo, and Pang}]{long2024learning}
Long, J.; Ren, J.; Shi, M.; Wang, Z.; Huang, T.; Luo, P.; and Pang, J. 2024.
\newblock Learning Humanoid Locomotion with Perceptive Internal Model.
\newblock \emph{arXiv preprint arXiv:2411.14386}.

\bibitem[{Malik, Masood, and Brem(2023)}]{malik2023intelligent}
Malik, A.~A.; Masood, T.; and Brem, A. 2023.
\newblock Intelligent humanoids in manufacturing to address worker shortage and skill gaps: Case of Tesla Optimus.
\newblock \emph{arXiv preprint arXiv:2304.04949}.

\bibitem[{Mende et~al.(2019)Mende, Scott, Van~Doorn, Grewal, and Shanks}]{mende2019service}
Mende, M.; Scott, M.~L.; Van~Doorn, J.; Grewal, D.; and Shanks, I. 2019.
\newblock Service robots rising: How humanoid robots influence service experiences and elicit compensatory consumer responses.
\newblock \emph{Journal of Marketing Research}, 56(4): 535--556.

\bibitem[{Moos et~al.(2022)Moos, Hansel, Abdulsamad, Stark, Clever, and Peters}]{moos2022robust}
Moos, J.; Hansel, K.; Abdulsamad, H.; Stark, S.; Clever, D.; and Peters, J. 2022.
\newblock Robust reinforcement learning: A review of foundations and recent advances.
\newblock \emph{Machine Learning and Knowledge Extraction}, 4(1): 276--315.

\bibitem[{Mukherjee et~al.(2022)Mukherjee, Baral, Pal, Chittipaka, Roy, and Alam}]{mukherjee2022humanoid}
Mukherjee, S.; Baral, M.~M.; Pal, S.~K.; Chittipaka, V.; Roy, R.; and Alam, K. 2022.
\newblock Humanoid robot in healthcare: a systematic review and future research directions.
\newblock In \emph{2022 International conference on machine learning, big data, cloud and parallel computing (COM-IT-CON)}, volume~1, 822--826. IEEE.

\bibitem[{Mysore et~al.(2021)Mysore, Mabsout, Mancuso, and Saenko}]{mysore2021regularizing}
Mysore, S.; Mabsout, B.; Mancuso, R.; and Saenko, K. 2021.
\newblock Regularizing action policies for smooth control with reinforcement learning.
\newblock In \emph{2021 IEEE International Conference on Robotics and Automation (ICRA)}, 1810--1816. IEEE.

\bibitem[{Peng et~al.(2018)Peng, Andrychowicz, Zaremba, and Abbeel}]{peng2018sim}
Peng, X.~B.; Andrychowicz, M.; Zaremba, W.; and Abbeel, P. 2018.
\newblock Sim-to-real transfer of robotic control with dynamics randomization.
\newblock In \emph{2018 IEEE international conference on robotics and automation (ICRA)}, 3803--3810. IEEE.

\bibitem[{Perolat et~al.(2015)Perolat, Scherrer, Piot, and Pietquin}]{perolat2015approximate}
Perolat, J.; Scherrer, B.; Piot, B.; and Pietquin, O. 2015.
\newblock Approximate dynamic programming for two-player zero-sum Markov games.
\newblock In \emph{International Conference on Machine Learning}, 1321--1329. PMLR.

\bibitem[{Radosavovic et~al.(2024)Radosavovic, Zhang, Shi, Rajasegaran, Kamat, Darrell, Sreenath, and Malik}]{radosavovic2024humanoid}
Radosavovic, I.; Zhang, B.; Shi, B.; Rajasegaran, J.; Kamat, S.; Darrell, T.; Sreenath, K.; and Malik, J. 2024.
\newblock Humanoid locomotion as next token prediction.
\newblock In \emph{The Thirty-eighth Annual Conference on Neural Information Processing Systems}.

\bibitem[{Ren et~al.(2025)Ren, Huang, Wang, Wang, Ben, Pang, and Luo}]{ren2025vb}
Ren, J.; Huang, T.; Wang, H.; Wang, Z.; Ben, Q.; Pang, J.; and Luo, P. 2025.
\newblock Vb-com: Learning vision-blind composite humanoid locomotion against deficient perception.
\newblock \emph{arXiv preprint arXiv:2502.14814}.

\bibitem[{Salvato et~al.(2021)Salvato, Fenu, Medvet, and Pellegrino}]{salvato2021crossing}
Salvato, E.; Fenu, G.; Medvet, E.; and Pellegrino, F.~A. 2021.
\newblock Crossing the reality gap: A survey on sim-to-real transferability of robot controllers in reinforcement learning.
\newblock \emph{IEEE Access}, 9: 153171--153187.

\bibitem[{Schulman et~al.(2017)Schulman, Wolski, Dhariwal, Radford, and Klimov}]{schulman2017proximal}
Schulman, J.; Wolski, F.; Dhariwal, P.; Radford, A.; and Klimov, O. 2017.
\newblock Proximal policy optimization algorithms.
\newblock \emph{arXiv preprint arXiv:1707.06347}.

\bibitem[{Shi et~al.(2024)Shi, Zhang, Miki, Lee, Hutter, and Coros}]{shi2024rethinking}
Shi, F.; Zhang, C.; Miki, T.; Lee, J.; Hutter, M.; and Coros, S. 2024.
\newblock Rethinking Robustness Assessment: Adversarial Attacks on Learning-based Quadrupedal Locomotion Controllers.
\newblock \emph{arXiv preprint arXiv:2405.12424}.

\bibitem[{Shi et~al.(2025)Shi, Liu, Wang, Lu, Schwertfeger, Sun, Bai, and Li}]{shi2025adversarial}
Shi, J.; Liu, X.; Wang, D.; Lu, O.; Schwertfeger, S.; Sun, F.; Bai, C.; and Li, X. 2025.
\newblock Adversarial Locomotion and Motion Imitation for Humanoid Policy Learning.
\newblock \emph{arXiv preprint arXiv:2504.14305}.

\bibitem[{Su et~al.(2024)Su, Huang, Ordo{\~n}ez-Apraez, Li, Li, Liao, Turrisi, Pontil, Semini, Wu et~al.}]{su2024leveraging}
Su, Z.; Huang, X.; Ordo{\~n}ez-Apraez, D.; Li, Y.; Li, Z.; Liao, Q.; Turrisi, G.; Pontil, M.; Semini, C.; Wu, Y.; et~al. 2024.
\newblock Leveraging symmetry in rl-based legged locomotion control.
\newblock In \emph{2024 IEEE/RSJ International Conference on Intelligent Robots and Systems (IROS)}, 6899--6906. IEEE.

\bibitem[{Sun et~al.(2025)Sun, Cao, Chen, Su, Liu, Xie, and Liu}]{sun2025learning}
Sun, W.; Cao, B.; Chen, L.; Su, Y.; Liu, Y.; Xie, Z.; and Liu, H. 2025.
\newblock Learning Perceptive Humanoid Locomotion over Challenging Terrain.
\newblock \emph{arXiv preprint arXiv:2503.00692}.

\bibitem[{Tang, Tan, and Harada(2020)}]{tang2020learning}
Tang, Y.; Tan, J.; and Harada, T. 2020.
\newblock Learning agile locomotion via adversarial training.
\newblock In \emph{2020 IEEE/RSJ International Conference On Intelligent Robots And Systems (IROS)}, 6098--6105. IEEE.

\bibitem[{Tessler, Efroni, and Mannor(2019)}]{tessler2019action}
Tessler, C.; Efroni, Y.; and Mannor, S. 2019.
\newblock Action robust reinforcement learning and applications in continuous control.
\newblock In \emph{International Conference on Machine Learning}, 6215--6224. PMLR.

\bibitem[{Tong, Liu, and Zhang(2024)}]{tong2024advancements}
Tong, Y.; Liu, H.; and Zhang, Z. 2024.
\newblock Advancements in humanoid robots: A comprehensive review and future prospects.
\newblock \emph{IEEE/CAA Journal of Automatica Sinica}, 11(2): 301--328.

\bibitem[{{Unitree G1}(2025)}]{example_website}
{Unitree G1}. 2025.
\newblock \url{https://www.unitree.com/cn/g1}.
\newblock [Online; accessed 25-June-2025].

\bibitem[{van Marum et~al.(2024)van Marum, Shrestha, Duan, Dugar, Dao, and Fern}]{van2024revisiting}
van Marum, B.; Shrestha, A.; Duan, H.; Dugar, P.; Dao, J.; and Fern, A. 2024.
\newblock Revisiting Reward Design and Evaluation for Robust Humanoid Standing and Walking.
\newblock \emph{arXiv preprint arXiv:2404.19173}.

\bibitem[{Wang, Liu, and Li(2020)}]{wang2020reinforcement}
Wang, J.; Liu, Y.; and Li, B. 2020.
\newblock Reinforcement learning with perturbed rewards.
\newblock In \emph{Proceedings of the AAAI conference on artificial intelligence}, volume~34, 6202--6209.

\bibitem[{Wei et~al.(2023)Wei, Wang, Xie, Wu, Xiong, and Zhu}]{wei2023learning}
Wei, W.; Wang, Z.; Xie, A.; Wu, J.; Xiong, R.; and Zhu, Q. 2023.
\newblock Learning Gait-conditioned Bipedal Locomotion with Motor Adaptation.
\newblock In \emph{2023 IEEE-RAS 22nd International Conference on Humanoid Robots (Humanoids)}, 1--7. IEEE.

\bibitem[{Yan et~al.(2024)Yan, Zhang, Zhang, Boedecker, and Burgard}]{yan2024learning}
Yan, S.; Zhang, B.; Zhang, Y.; Boedecker, J.; and Burgard, W. 2024.
\newblock Learning continuous control with geometric regularity from robot intrinsic symmetry.
\newblock In \emph{2024 IEEE International Conference on Robotics and Automation (ICRA)}, 49--55. IEEE.

\bibitem[{Ze et~al.(2025)Ze, Chen, Ara{\~A}{\v{s}}jo, Cao, Peng, Wu, and Liu}]{ze2025twist}
Ze, Y.; Chen, Z.; Ara{\~A}{\v{s}}jo, J.~P.; Cao, Z.-a.; Peng, X.~B.; Wu, J.; and Liu, C.~K. 2025.
\newblock TWIST: Teleoperated Whole-Body Imitation System.
\newblock \emph{arXiv preprint arXiv:2505.02833}.

\bibitem[{Zhang et~al.(2021)Zhang, Chen, Boning, and Hsieh}]{zhang2021robust}
Zhang, H.; Chen, H.; Boning, D.; and Hsieh, C.-J. 2021.
\newblock Robust reinforcement learning on state observations with learned optimal adversary.
\newblock \emph{arXiv preprint arXiv:2101.08452}.

\bibitem[{Zhang et~al.(2024)Zhang, Cui, Yan, Sun, Duan, Han, Zhao, Zhang, Guo, Zhang et~al.}]{zhang2024whole}
Zhang, Q.; Cui, P.; Yan, D.; Sun, J.; Duan, Y.; Han, G.; Zhao, W.; Zhang, W.; Guo, Y.; Zhang, A.; et~al. 2024.
\newblock Whole-body humanoid robot locomotion with human reference.
\newblock In \emph{2024 IEEE/RSJ International Conference on Intelligent Robots and Systems (IROS)}, 11225--11231. IEEE.

\bibitem[{Zhang, Nie, and Gao(2024)}]{zhang2024robust}
Zhang, Y.; Nie, B.; and Gao, Y. 2024.
\newblock Robust Locomotion Policy with Adaptive Lipschitz Constraint for Legged Robots.
\newblock \emph{IEEE Robotics and Automation Letters}.

\bibitem[{Zhuang, Yao, and Zhao(2024)}]{zhuang2024humanoid}
Zhuang, Z.; Yao, S.; and Zhao, H. 2024.
\newblock Humanoid Parkour Learning.
\newblock \emph{arXiv preprint arXiv:2406.10759}.

\end{thebibliography}



\clearpage 

\section{Appendix}

\subsection{Deployment Setup}
Simulation training is implemented on a computing platform built with NVIDIA GeForce RTX $4090$ GPU devices, based on the IsaacGym and LeggedGym frameworks to realize our proposed method.
For deployment in the real robot, the Unitree G1 humanoid robot is adopted, with its Jetson Orin NX on-board serving as the main control unit, responsible for real-time motion control and communication.
All $29$ degrees of freedom of the G1 robot are activated, giving it flexible whole-body movement capabilities.
The environment perception module processes the real-time conversion from LiDAR point clouds to elevation maps through an external laptop (equipped with an NVIDIA GeForce RTX $3090$ GPU device), ensuring a stable inference frequency.

\subsection{Learning Robust Perceptive Locomotion}\label{Implementation1}
In Isaac Gym simulation, we can directly obtain the information of the local elevation map in the robot's base Z-aligned coordinate system.
In real robot deployment, the Livox Mid-$360$ LiDAR mounted on the head of the G1 humanoid robot is used to run Fast-LIO to acquire denoised point clouds and odometry, which are then sent to the elevation mapping algorithm to compute the local elevation map in the LiDAR coordinate system.
Using the IMU on the robot itself and the three-degree-of-freedom joint angles of the waist, the elevation map aligned with that in the simulation is obtained through coordinate transformation to the Base coordinate system.
In addition, due to the absorption of LiDAR light by certain materials in the environment, there are outliers in the elevation map.
We improve stability by obtaining the relative elevation map information centered on the base through local averaging.
To keep the simulation as consistent as possible with the real robot, the update frequency of elevation map information in the simulation is set to $10Hz$, and the policy inference frequency is $50Hz$.

We train a velocity-command-tracking locomotion policy for the Unitree G1 humanoid robot ($29$ degrees of freedom), utilizing a LiDAR-based elevation map to perceive unstructured terrain.
An asymmetric actor-critic structure is employed to estimate privileged information from historical observations, which is then fed into the policy network, enabling single-stage policy learning without distillation.

\begin{figure}[t]
\centering  
\subfigure[]{
\label{sfig1-1}
\includegraphics[width=4.0cm]{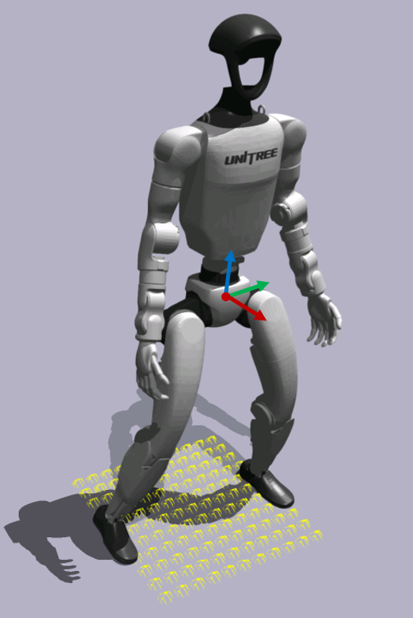}}
\subfigure[]{
\label{sfig1-2}
\includegraphics[width=4.0cm]{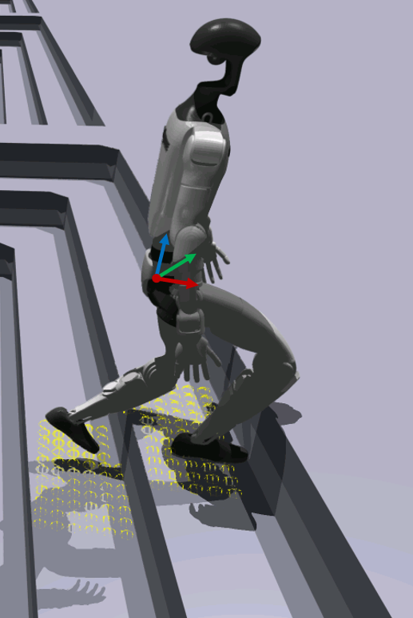}}
\caption{Visualization of the robot's elevation map in the base coordinate system during Isaac Gym simulation.
(a) Elevation map on flat ground.
(b) Elevation map on stairs.}
\label{sfig1}
\end{figure}

\begin{table}[!ht]
    \centering
    \begin{tabular}{ll} 
        \toprule
        \textbf{Proprioceptive States} & \textbf{Dim} \\
        
        \midrule
    Base angular velocity $\boldsymbol{\omega}_{t}$ & 3 \\
    Projected gravity $\boldsymbol{g}_{t}$  & 3 \\
    Velocity commands $\boldsymbol{c}_{t}$ & 3 \\
    Joint positions $\boldsymbol{q}_{t}$ & 29 \\
    Joint velocities $\boldsymbol{v}_{t}$ & 29 \\
    Previous action $\boldsymbol{a}_{t-1}$ & 29 \\
    Height map $\boldsymbol{h}_{t}$ & 121 \\

    \midrule 
        \textbf{Privileged Information} & \textbf{Dim} \\
    Base linear velocity $\boldsymbol{v}_{t}$ & 3\\
    Base mass  & 1 \\
    Friction & 1\\
    Motor strength & 29 \\
    Motor offset & 29 \\
    Motor Kp & 29 \\
    Motor Kd & 29 \\
    
    \bottomrule
    \end{tabular}
    \caption{Proprioceptive and privileged information in perceptive locomotion policy training.}
    \label{stable1}
\end{table}

\begin{table*}[t]
    \centering
    \begin{tabular}{llcc} 
        \toprule
        \textbf{Reward terms} & \textbf{Expression} & \textbf{Weight} \\
        
        \midrule

    Linear velocity tracking & $\exp(-\Vert v-v^{cmd}\Vert/0.25)$ & $2.0$ \\
    Angular velocity tracking & $\exp(-\Vert \omega_{yaw}-\omega_{yaw}^{cmd}\Vert/0.25)$ & $1.0$ \\
    Orientation & $g_{x}^{2}+g_{y}^{2}$ & $-20.0$ \\

    Dof velocity & $\sum_{j\in joints}\vert \dot{q}_{j} \vert^{2}$  & $-1e^{-6}$  \\
    Dof acceleration & $\sum_{j\in joints}\vert \ddot{q}_{j} \vert^{2}$ & $-2.5e^{-7}$ \\
    
    Alive & $\mathbf{1}_{alive}$ & $0.5$ \\
    Penalize linear velocity on z axis & $v_{z}^{2}$ & $-2.0$ \\
    Penalize angular velocity on xy axis & $\sum(\omega_{x}^2 + \omega_{y}^2)$ & $-0.1$ \\
    
    Base height & $(h_{b}-0.7)^{2}$ & $-100.0$ \\
    
    Action rate & $\sum(\text{a}_{t} - \text{a}_{t-1})^2$ & $-1e^{-4}$ \\
    Action smoothness & $\sum(\text{a}_{t} - 2\text{a}_{t-1} + \text{a}_{t-2})^2$ & $-1e^{-4}$ \\

    Torques & $\sum_{j\in joints}\tau_{j}^{2}$ & $-6e^{-7}$ \\

    Hip yaw dof & $\sum_{j\in hip\,yaw} \vert q_{j}\vert^{2}$ & $-1.0$ \\
    Waist dof & $\sum_{j\in waist} \vert q_{j}\vert^{2}$ & $-10.0$ \\
    Arm dof & $\sum_{j\in arm}\vert q_{j}\vert^{2}$ & $-0.5$ \\
    
    Feet clearance & $\sum(\Vert h_{z,foot}-0.25\Vert^{2} \times v_{foot,air})$ & $-0.1$ \\
    Contact force & $\mathbf{1}\{\vert F_{i}\geq400\vert\}\times(\vert F_{i}\vert - 400)$ & $-1e^{-4}$ \\
    Collision & $\sum_{i\in contact} \mathbf{1}\{\Vert F_{i}\Vert>0.1\}$ & $-1.0$ \\
        \bottomrule
    \end{tabular}
    \caption{The description of the reward function utilized in perceptive locomotion policy training.}
    \label{stable2}
\end{table*}

\subsubsection{Observation and Action Spaces}
We sample $121$ elevation points within the Z-axis-aligned robot frame in the simulation, as shown in Fig.~\ref{sfig1}.
These points are distributed across a $1.0\,m\times1.0\,m$ square centered on the robot, and their deviations from the mean value are used as the perceptive input $\boldsymbol{h}_{t}$ for the locomotion policy.
The locomotion policy’s observation $\boldsymbol{o}_{t}=[\boldsymbol{c}_{t},\boldsymbol{\omega}_{t},\boldsymbol{g}_{t},\boldsymbol{q}_{t},\dot{\boldsymbol{q}}_{t},\boldsymbol{a}_{t-1},\boldsymbol{h}_{t}]$ contains velocity command $\boldsymbol{c}_{t}=[v_{x}^{c},v_{y}^{c},\omega_{yaw}^{c}]$, base angular velocity $\boldsymbol{\omega}_{t}$, gravity projection $\boldsymbol{g}_{t}$, joint position $\boldsymbol{q}_{t}$, joint velocity $\dot{\boldsymbol{q}}_{t}$, last timestep action $\boldsymbol{a}_{t-1}$, and perceptive information $\boldsymbol{h}_{t}$.
The critic is allowed to access additional privileged information such as base linear velocity $v_{t}$ and other environmental parameters.
All proprioceptive states and privileged information can be found in Table~\ref{stable1}.
The locomotion policy outputs target joint positions, which are subsequently converted to joint torques through PD controllers.

\subsubsection{Reward Function}
The reward function for training the perceptive locomotion policy is shown in Table~\ref{stable2}, which mainly consists of velocity tracking rewards and regularization terms to make the policy output actions smoother and avoid constraint violations.

\subsubsection{Symmetry Regularization}
In order to achieve a more symmetrical and coordinated human-like movement pattern, we incorporate a symmetry constraint loss $\mathcal{L}_{sym}$ into the policy optimization process:
\begin{equation}\label{seq1}
\mathcal{L}_{sym}=MSE(\pi^{m}(s_{t}),M^{a}\cdot\pi^{m}(M^{s}\cdot s_{t})),
\end{equation}
where $M^{s}$ and $M^{a}$ represent the transformation matrices that mirror the state $s_{t}$ and action $a_{t}$ relative to the X-Z plane, respectively.

\begin{table*}[t]
    \centering
    \begin{tabular}{llcc} 
        \toprule
        \textbf{Attack terms} & \textbf{Dim} & \textbf{Expression} & \textbf{Threshold} \\
        
        \midrule
    Proprioceptive attack $\delta_{s1,\boldsymbol{\omega}}$ & $3$ & $\boldsymbol{\omega}_{t}+\delta_{s1,\boldsymbol{\omega}}$ & $[-0.2,0.2]$ \\
    Proprioceptive attack $\delta_{s1,\boldsymbol{g}}$ & $3$ & $\boldsymbol{g}_{t}+\delta_{s1,\boldsymbol{g}_{t}}$ & $[-0.05,0.05]$ \\
    Proprioceptive attack $\delta_{s1,\boldsymbol{q}}$ & $29$ & $\boldsymbol{q}_{t}+\delta_{s1,\boldsymbol{q}}$ & $[-0.015,0.015]$ \\
    Proprioceptive attack $\delta_{s1,\dot{\boldsymbol{q}}}$ & $29$ & $\dot{\boldsymbol{q}}_{t}+\delta_{s1,\dot{\boldsymbol{q}}}$ & $[-0.5,0.5]$ \\
    Elevation map attack $\delta_{s2}$ & $121$ & $\boldsymbol{h}_{t}\times\delta_{s2}$ & $[0.8,1.2]$\\
    Elevation map attack $\delta_{s3}$ & $121$ & $(\boldsymbol{h}_{t}\times\delta_{s2})+\delta_{s3}$ & $[-0.025,0.025]$ \\
    Action attack $\delta_{a1}$ & $29$ & $\boldsymbol{a}_{t}+\delta_{a1}$ & $[-0.02,0.02]$ \\
    Kp attack $\delta_{a2}$ & $29$ & $Kp\times\delta_{a2}$ & $[0.95,1.05]$ \\
    Kd attack $\delta_{a3}$ & $29$ & $Kd\times\delta_{a2}$ & $[0.95,1.05]$ \\
    Torque attack $\delta_{a4}$ & $29$ & $\boldsymbol{\tau}\times\delta_{a4}$ & $[0.95, 1.05]$\\
    Binary attack variable $b$ & $1$ & $[\delta_{s1},\delta_{s2},\delta_{s3},\delta_{a1},\delta_{a2},\delta_{a3},\delta_{a4}]\times b$ & \{0,1\}\\
        \bottomrule
    \end{tabular}
    \caption{Attack space and attack threshold of selective attack policy.}
    \label{stable3}
\end{table*}

\begin{table}[!ht]
    \centering
    \begin{tabular}{llcc} 
        \toprule
        \textbf{Reward terms} & \textbf{Expression} & \textbf{Weight} \\
        
        \midrule

     Robot falling & $\vmathbb{1}$ & $1.0$ \\
     Orientation  & $(\theta_{r}^{2}+\theta_{p}^{2})$ & $0.5$ \\
     Base angular velocity & $(\omega_{x}^{2}+\omega_{y}^{2})$ & $0.1$ \\
     Base height & $v_{z}^{2}$ & $0.5$ \\
     Attack Budget & $\sum_{i\in\{0,\ldots,T\}}b_{i}$ & $0.1$\\
        \bottomrule
    \end{tabular}
    \caption{Reward function for selective attack policy.}
    \label{stable4}
\end{table}

\subsubsection{Selective Attack Policy}
To effectively attack the locomotion policy, the SAP's observation space comprises both the locomotion policy's observations $o_{t}$ and all privileged information.
The action space of SAP is defined as:
\begin{equation}\label{seq2}
a^{adv}=[\delta_{s1},\delta_{s2},\delta_{s3},\delta_{a1},\delta_{a2},\delta_{a3},\delta_{a4}]\times b,
\end{equation}
which including additive attack $\delta_{s1}$ to the proprioceptive information, multiplicative attack $\delta_{s2}$ and additive attack $\delta_{s3}$ to the perceptive information $h_{t}$ (formulated as $h_{t}^{attack}=\delta_{s2}\times h_{t}+\delta_{s3}$), additive attack $\delta_{a1}$ to the policy output action $a_{t}$, the $K_{P}$ and $K_{D}$ of the PD controller apply multiplicative attack $\delta_{a2}$ and $\delta_{a3}$, respectively, multiplicative attack $\delta_{a4}$ to joint torques, and binary attack variable $b\in\{0,1\}$.
All attack parameters and attack thresholds are shown in Table~\ref{stable3}.
It should be noted that we maintain consistency between our attack space and the perturbation space of previous work using DR, enabling direct experimental comparison to validate our method's effectiveness.
The adversarial reward function focuses exclusively on the robot's stability metrics, which can be formally expressed as:
\begin{equation}\label{seq3}
R^{adv}=c_{1}\cdot\vmathbb{1}_{fall}+c_{2}\cdot(\theta_{r}^{2}+\theta_{p}^{2}) + c_{3}\cdot(\omega_{x}^{2}+\omega_{y}^{2})+c_{4}\cdot v_{z}^{2},
\end{equation}
where $\vmathbb{1}_{fall}$ denotes robot falling, $\theta_{r}$ and $\theta_{p}$ represent the base's roll and pitch angles, respectively, $\omega$ indicates base angular velocity, and $v$ indicates base linear velocity.
The coefficients $c_{i},i\in\{1,2,3,4\}$ are weighting parameters, as shown in Table~\ref{stable4}.
These terms collectively guide the adversarial attack policy to effectively attack the robot in unsafe states.
Furthermore, to simulate more realistic perturbation characteristics in practical scenarios, we impose Lipschitz regularization on adversarial policy optimization.
By incorporating the infinite norm of the policy network parameters as a regularization loss term $\mathcal{L}_{lips}=\prod_{i=1}\Vert\theta_{adv}^{i}\Vert_{\infty}$ to constrain the Lipschitz constant of the network, thereby ensuring smoother adversarial attacks.

\subsubsection{Training Details}
We alternately optimize the perceptive locomotion policy and the selective attack policy, with all networks utilizing Multi-Layer Perceptron (MLP).
Detailed network structure parameters and training hyperparameters are presented in Table~\ref{stable5}.

\begin{table}[t]
    \centering
    \begin{tabular}{ll} 
        \toprule
        \textbf{Hyperparameter} & \textbf{Value} \\
        \midrule 

        Optimizer & Adam \\
        PPO clip range & $0.2$ \\
        GAE $\lambda$ & $0.95$ \\
        Learning rate & $1e^{-4}$ \\
        Reward discount factor & $0.99$ \\
        Entropy coefficient & $1e^{-3}$ \\
        Value loss coefficient & $1.0$ \\
        Symmetry coefficient & $20.0$ \\
        Max gradient norm & $1.0$ \\
        Lipschitz coefficient & $1e^{-3}$ \\
        Total iterations $N_{iter}$ & $6e^{3}$ \\
        Attack policy iteration $N_{adv}$ & $1e^{2}$ \\
        Motion policy iteration $N_{m}$ & $1e^{2}$ \\
        State encoder MLP size &  $[512,256,128]$ \\
        State decoder MLP size &  $[128,256,512]$ \\
        Motion actor MLP size &  $[512,256,128]$ \\
        Motion value MLP size &  $[512,256,128]$ \\
        Attacker actor MLP size & $[512,256,128]$ \\
        Attacker value 1 MLP size & $[512,256,128]$ \\
        Attacker value 2 MLP size & $[512,256,128]$ \\
    \bottomrule
    \end{tabular}
    \caption{Hyperparameters related to perceptive locomotion policy training.}
    \label{stable5}
\end{table}

\begin{table*}[t]
    \centering
    \begin{tabular}{llcc} 
        \toprule
        \textbf{Reward terms} & \textbf{Expression} & \textbf{Weight} \\
        
        \midrule

    Base pos tracking & $\exp(-\Vert\Delta p - \Delta p^{cmd}\Vert/ 0.01^{2})$ & 5.0 \\
    Base orientation tracking & $\exp(-\Vert\Delta Q - \Delta Q^{cmd}\Vert/0.1^{2})$ & 1.0 \\
    Dof pos tracking & $\exp(-(\sum_{j\in joints}\vert q_{j}-q_{j}^{cmd} \vert^{2})/2.0^{2})$ & 10.0 \\
    Keypoints tracking & $\exp(-(\sum_{j\in joints}\vert \Delta p^{k} - p^{k,cmd} \vert^{2})/0.5^{2})$  & 1.0 \\

    Dof velocity & $\sum_{j\in joints}\vert \dot{q}_{j} \vert^{2}$  & $-1e^{-6}$  \\
    Dof acceleration & $\sum_{j\in joints}\vert \ddot{q}_{j} \vert^{2}$ & $-2.5e^{-7}$ \\
    
    Alive & $\mathbf{1}_{alive}$ & $0.5$ \\
    
    Action rate & $\sum(\text{a}_{t} - \text{a}_{t-1})^2$ & $-1e^{-3}$ \\
    Action smoothness & $\sum(\text{a}_{t} - 2\text{a}_{t-1} + \text{a}_{t-2})^2$ & $-1e^{-4}$ \\

    Torques & $\sum_{j\in joints}\tau_{j}^{2}$ & $-6e^{-5}$ \\

    Contact force & $\mathbf{1}\{\vert F_{i}\geq400\vert\}\times(\vert F_{i}\vert - 400)$ & $-1e^{-1}$ \\
    Collision & $\sum_{i\in contact} \mathbf{1}\{\Vert F_{i}\Vert>0.1\}$ & $-1.0$ \\
        \bottomrule
    \end{tabular}
    \caption{Reward function for whole-body control policy training.}
    \label{stable8}
\end{table*}

\subsection{Learning Robust Whole-Body Control}
We formulate the whole-body control problem as a target trajectory imitation learning task, where the imitation policy is trained to track the whole-body trajectory.
The motion trajectories are derived from the open-source LAFAN1 dataset, remapped to the Unitree G1 humanoid robot, and upsampled to a frequency of $50Hz$.
Each trajectory includes the base position, the base quaternion, and all joint angles.
We replay these trajectories in simulation and collect keypoints of robot bodies to augment the reference trajectory information.
The detailed information for each frame of the reference trajectory is shown in Table~\ref{stable6}.

\begin{table}[t]
    \centering
    \begin{tabular}{ll} 
        \toprule
        \textbf{Trajectory information} & \textbf{Dim} \\
        \midrule 
        Base position & $3$ \\
        Base quaternion & $4$ \\
        Joint angle & $29$ \\
        Keypoints & $203$ \\
    \bottomrule
    \end{tabular}
    \caption{The information of each frame in the reference trajectory.}
    \label{stable6}
\end{table}

\subsubsection{Observation and Action Spaces}
The imitation policy takes $[o_{t},c^{traj}_{t}]$ as input and the output action $a_{t}$ is converted into torque by the PD controller.
The $c^{traj}_{t}$ denotes the target trajectory frame at timestep $t$, represented in the robot's local frame.
The observations of the whole-body control policy are shown in Table~\ref{stable7}.

\begin{table}[t]
    \centering
    \begin{tabular}{ll} 
        \toprule
        \textbf{Observations} & \textbf{Dim} \\
        \midrule 
    Base angular velocity $\boldsymbol{\omega}_{t}$ & 3 \\
    Projected gravity $\boldsymbol{g}_{t}$  & 3 \\
    Velocity commands $\boldsymbol{c}_{t}$ & 3 \\
    Joint positions $\boldsymbol{q}_{t}$ & 29 \\
    Joint velocities $\boldsymbol{v}_{t}$ & 29 \\
    Previous action $\boldsymbol{a}_{t-1}$ & 29 \\
    Target base pos delta $p_{t}-p_{t-1}$  & 3 \\
    Target base quat delta $Q_{t}\otimes Q_{t-1}^{-1}$ & 4 \\
    Target dof pos $q_{target}-q_{default}$ & 29 \\
    Target Keypoints delta $p_{t}^{k}-p_{t-1}^{k}$ & 203 \\
    \bottomrule
    \end{tabular}
    \caption{Observation of whole-body control policy.}
    \label{stable7}
\end{table}

\subsubsection{Reward Function}
The reward function, following previous work, primarily consists of task-specific rewards and several regularization rewards aimed at enhancing stability. Detailed designs of the reward function are presented in Table~\ref{stable8}.

\subsubsection{Training Details}
Similarly to perceptive locomotion, the symmetry constraint~\eqref{seq1} is integrated into the WBC policy training framework.
In the design of adversarial policy against the whole-body imitation policy, we exclude perceptive information attacks $[\delta_{s2},\delta_{s3}]$ from the attack space designed in perceptive locomotion, while all other training components (including the observation space and reward function) remain the same.
All hyperparameters for training are shown in Table~\ref{stable9}.

\begin{table}[t]
    \centering
    \begin{tabular}{ll} 
        \toprule
        \textbf{Hyperparameter} & \textbf{Value} \\
        \midrule 
        Optimizer & Adam \\
        PPO clip range & $0.2$ \\
        GAE $\lambda$ & $0.95$ \\
        Learning rate & $1e^{-4}$ \\
        Reward discount factor & $0.99$ \\
        Entropy coefficient & $1e^{-3}$ \\
        Value loss coefficient & $1.0$ \\
        Symmetry coefficient & $20.0$ \\
        Max gradient norm & $1.0$ \\
        Lipschitz coefficient & $1e^{-3}$ \\
        Total iterations $N_{iter}$ & $1e^{4}$ \\
        Attack policy iteration $N_{adv}$ & $1e^{2}$ \\
        Motion policy iteration $N_{m}$ & $1e^{2}$ \\
        Motion actor MLP size &  $[512,256,128]$ \\
        Motion value MLP size &  $[1024, 512,256,128]$ \\
        Attacker actor MLP size & $[512,256,128]$ \\
        Attacker value 1 MLP size & $[512,256,128]$ \\
        Attacker value 2 MLP size & $[512,256,128]$ \\
    \bottomrule
    \end{tabular}
    \caption{Hyperparameters related to whole-body control policy training.}
    \label{stable9}
\end{table}

\subsection{Experimental Details}
\subsubsection{Evaluation metrics}
We use several metrics to evaluate the performance of motion policies. The implementation details of each metric are as follows:
\begin{itemize}
\item The mean linear velocity error $E_{vel}(m/s)$ is used to evaluate the linear velocity tracking performance of the motion policy.
Under the influence of perturbations, a smaller mean linear velocity error indicates better task robustness.
\begin{equation}
    E_{vel} = \mathbb{E}_{\pi}[\Vert v_{x,y} - v_{x,y}^{cmd}\Vert]
\end{equation}

\item The mean angular velocity error $E_{ang}(rad/s)$ is used to evaluate the tracking performance of the steering angular velocity $\omega_{yaw}$ for the motion policy.
\begin{equation}
    E_{ang} = \mathbb{E}_{\pi}[\Vert \omega_{yaw} - \omega_{yaw}^{cmd}\Vert]
\end{equation}

\item The gravitational projection component $E_{g}(m/s^{2})$ represents the x-y components of the unit gravitational vector, describing the base's inclination tendency during the robot's motion, and is used to evaluate the stability of the robot.
\begin{equation}
    E_{g} = \mathbb{E}_{\pi}[g_{x}^{2}+g_{y}^{2}]
\end{equation}

\item The Mean Per Keybody Position Error (MPKPE), $E_{mpkpe}^{upper}(m)$ for the upper body and $E_{mpkpe}^{lower}(m)$ for the lower body, is used to evaluate the keypoint trajectory tracking accuracy of the whole-body control policy. 
The robot dynamically adjusts the foot positions via its lower body to maintain balance and enable motion, making keypoint tracking of the lower body more challenging.
\begin{equation}
\begin{split}
    & E_{mpkpe}^{upper} = \mathbb{E}_{\pi}[\frac{1}{N_{upper}}\sum_{i\in N_{upper}}\Vert p_{i}^{k}-p_{i}^{k,cmd}\Vert] \\
    & E_{mpkpe}^{lower} = \mathbb{E}_{\pi}[\frac{1}{N_{lower}}\sum_{i\in N_{lower}}\Vert p_{i}^{k}-p_{i}^{k,cmd}\Vert] \\
\end{split}
\end{equation}

\item The Mean Per Joint Position Error (MPJPE), $E_{mpjpe}^{upper}(rad)$ for the upper body and $E_{mpjpe}^{lower}(rad)$ for the lower body, is used to evaluate the joint position tracking accuracy.
\begin{equation}
\begin{split}
    & E_{mpjpe}^{upper} = \mathbb{E}_{\pi}[\frac{1}{N_{upper}}\sum_{i\in N_{upper}}\Vert q_{i}-q_{i}^{cmd}\Vert] \\
    & E_{mpjpe}^{lower} = \mathbb{E}_{\pi}[\frac{1}{N_{lower}}\sum_{i\in N_{lower}}\Vert q_{i}-q_{i}^{cmd}\Vert] \\
\end{split}
\end{equation}

\item The success rate $R_{sr}(\%)$ for robot survival in complex terrain traversal and whole-body trajectory tracking.
\begin{equation}
    R_{sr} = \mathbb{E}_{\pi}[\frac{N_{success}}{N_{total}}\times100\%]
\end{equation}
\end{itemize}

\end{document}